\documentclass[11pt]{article}

\usepackage[english]{babel}
\usepackage{graphicx}
\usepackage{natbib}
\usepackage{amsfonts,amssymb,amsmath,amscd,amsthm,latexsym}
\usepackage{fullpage}
\usepackage{algorithm,algorithmic,nicefrac}
\usepackage{authblk}
\usepackage{multicol}

\newcommand{\bx}{\mathbf{x}}

\newcommand{\by}{\mathbf{y}}

\newcommand{\esp}{\mathbb{E}}

\newcommand{\prob}{\mathbb P}

\newcommand{\knn}{$k$-nn}

\makeatletter
\newenvironment{tablehere}
  {\def\@captype{table}}
  {}

\newenvironment{figurehere}
  {\def\@captype{figure}}
  {}
  
\newenvironment{algohere}
  {\def\@captype{algorithm}}
  {}
\makeatother

\begin{document}

\title{Reliable ABC model choice via random forests}

\author[1,2,6]{Pierre Pudlo}

\author[1,2,6]{Jean-Michel Marin\thanks{jean-michel.marin@umontpellier.fr}}

\author[3]{Arnaud Estoup}

\author[3]{Jean-Marie Cornuet}

\author[3]{Mathieu Gauthier}

\author[4,5]{Christian P. Robert}

\affil[1]{Universit\'e de Montpellier, IMAG, Montpellier, France}

\affil[2]{Institut de Biologie Computationnelle (IBC), Montpellier, France}

\affil[3]{CBGP, INRA, Montpellier, France}

\affil[4]{Universit\'e Paris Dauphine, CEREMADE, Paris, France}

\affil[5]{University of Warwick, Coventry, UK} 

\affil[6]{These authors contributed equally to this work} 

\maketitle

\begin{abstract}

  Approximate Bayesian computation (ABC) methods provide
  an elaborate approach to Bayesian inference on complex models,
  including model choice. Both theoretical arguments and simulation
  experiments indicate, however, that model posterior probabilities
  may be poorly evaluated by standard ABC techniques.

  We propose a novel approach based on a
  machine learning tool named random forests to conduct selection
  among the highly complex models covered by ABC algorithms. We thus
  modify the way Bayesian model selection is both understood
  and operated, in that we rephrase the inferential goal as a classification
  problem, first predicting the model that best fits the data with random forests
  and postponing the approximation of the posterior probability of the predicted MAP
  for a second stage also relying on random forests. Compared with earlier
   implementations of ABC model choice, the
  ABC random forest approach offers several potential improvements: \textit{(i)}
  it often has a larger discriminative power among the competing models,
  \textit{(ii)} it is more robust against the number and choice of statistics
  summarizing the data, \textit{(iii)} the computing effort is
  drastically reduced (with a gain in computation efficiency
  of at least fifty), and \textit{(iv)} it includes an
  approximation of the posterior probability of the selected model.
  The call to random forests will undoubtedly extend the range of
  size of datasets and complexity of models that ABC can handle. We
  illustrate the power of this novel methodology by
  analyzing controlled experiments as well as genuine population
  genetics datasets.
  
  The proposed methodologies are implemented in the R package {\sf abcrf} available on the CRAN.
  
  \vspace{0.5cm} \noindent \textbf{Keywords:} Approximate Bayesian Computation, model selection, summary
statistics, $k$-nearest neighbors, likelihood-free methods, random forests

\end{abstract}


\section{Introduction}

Approximate Bayesian Computation (ABC) represents an elaborate statistical approach to model-based inference
in a Bayesian setting in which model likelihoods are difficult to calculate (due to the complexity of the models
considered). Since its introduction in population genetics \citep{tavare:balding:griffith:donnelly:1997,
pritchard:seielstad:perez:feldman:1999,beaumont:zhang:balding:2002},
the method has found an ever increasing range of applications covering diverse types
of complex models in various scientific fields \citep[see, e.g.,][]{beaumont:2008, toni:etal:2009,
beaumont:2010,csillery:blum:gaggiotti:francois:2010,theunert:etal:2012,chan:etal:2014,arenas:etal:2015}. 
The principle of ABC is to conduct Bayesian inference on a
dataset through comparisons with numerous simulated datasets. However, it suffers from two major
difficulties. First, to ensure reliability of the method, the number of simulations is large; hence,
it proves difficult to apply ABC for large datasets (e.g., in population genomics where tens to
hundred thousand markers are commonly genotyped). Second, calibration has always been a critical
step in ABC implementation \citep{marin:pudlo:robert:ryder:2012,blum:nunes:prangle:sisson:2013}. More
specifically, the major feature in this calibration process involves selecting a vector of summary
statistics that quantifies the difference between the observed data and the simulated data. The
construction of this vector is therefore paramount and examples abound about poor performances of
ABC model choice algorithms related with specific choices of those statistics
\citep{didelot:everitt:johansen:lawson:2011,robert:cornuet:marin:pillai:2011,marin:pillai:robert:rousseau:2014},
even though there also are instances of successful implementations.

We advocate a drastic modification in the way ABC model selection is conducted: we propose both to step away from
selecting the most probable model from estimated posterior probabilities, and to reconsider the very problem of
constructing efficient summary statistics. First, given an arbitrary pool of available statistics, we now completely
bypass selecting among those. This new perspective directly proceeds from machine learning methodology.
Second, we postpone the approximation of model posterior probabilities to a second stage, as we deem the standard
numerical ABC approximations of such probabilities fundamentally untrustworthy. We instead advocate selecting
the posterior most
probable model by constructing a (machine learning) classifier from simulations from the prior predictive
distribution (or other distributions in more advanced versions of ABC), 
known as the ABC {\em reference table}. The statistical technique of random forests (RF)
\citep{breiman:2001} represents a trustworthy machine learning tool well adapted to complex settings as is typical for
ABC treatments.  Once the classifier is constructed and applied to the actual data, an approximation of the posterior
probability of the resulting model can be produced through a secondary random forest that regresses the selection error
over the available summary statistics.  We show here how RF improves upon existing classification methods in
significantly reducing both the classification error and the computational expense. After presenting theoretical
arguments, we illustrate the power of the ABC-RF methodology by analyzing controlled experiments as well as genuine
population genetics datasets.

\section{Materials and methods}

Bayesian model choice \citep{berger:1985,robert:2001} compares the fit of $M$ models to an observed dataset
$\bx^0$. It relies on a hierarchical modelling, setting first prior probabilities $\pi(m)$ on model indices
$m\in\{1,\ldots,M\}$ and then prior distributions $\pi(\theta|m)$ on the parameter $\theta$ of each model,
characterized by a likelihood function $f(\bx|m,\theta)$. Inferences and decisions are based on
the posterior probabilities of each model $\pi(m|\bx^0)$.

\subsection{ABC algorithms for model choice}

While we cannot cover in much details the principles of Approximate Bayesian computation (ABC), let us recall here that
ABC was introduced in \cite{tavare:balding:griffith:donnelly:1997} and \cite{pritchard:seielstad:perez:feldman:1999} for
solving intractable likelihood issues in population genetics. The reader is referred to, e.g., \cite{beaumont:2008},
\cite{toni:etal:2009}, \cite{beaumont:2010}, \cite{csillery:blum:gaggiotti:francois:2010} and
\cite{marin:pudlo:robert:ryder:2012} for thorough reviews on this approximation method. The fundamental principle at
work in ABC is that the value of the intractable likelihood function $f(\bx^0|\theta)$ at the observed data $\bx^0$ and
for a current parameter $\theta$ can be evaluated by the proximity between $\bx^0$ and pseudo-data $\bx(\theta)$
simulated from $f(\bx|\theta)$. In discrete settings, the indicator $\mathbb{I}(\bx(\theta)=\bx^0)$ is an unbiased
estimator of $f(\bx^0|\theta)$ \citep{rubin:1984}. For realistic settings, the equality constraint is replaced with a
tolerance region $\mathbb{I}(d(\bx(\theta),\bx^0)\le\epsilon)$, where $d(\bx^0,\bx)$ is a measure of divergence between
the two vectors and $\epsilon>0$ is a tolerance value. The implementation of this principle is straightforward: the ABC
algorithm produces a large number of pairs $(\theta,\bx)$ from the prior predictive, a collection called the {\em
reference table}, and extracts from the table the pairs $(\theta,\bx)$ for which $d(\bx(\theta),\bx^0)\le\epsilon$.

To approximate posterior probabilities of competing models, ABC methods
\citep{grelaud:marin:robert:rodolphe:tally:2009} compare observed data with a massive collection of pseudo-data,
generated from the prior predictive distribution in the most standard versions of ABC; the comparison proceeds via a
normalized Euclidean distance on a
vector of statistics $S(\bx)$ computed for both observed and simulated data. Standard ABC estimates posterior
probabilities $\pi(m|\bx^0)$ at stage (B) of Algorithm~\ref{algo:general} below as the frequencies of those models
within the $k$ nearest-to-$\bx^0$ simulations, proximity being defined by the distance between $S(\bx^0)$ and the
simulated $S(\bx)$'s.

Selecting a model means choosing the model with the highest frequency in the sample of size $k$
produced by ABC, such frequencies being approximations to posterior probabilities of models.  We
stress that this solution means resorting to a $k$-nearest neighbor (\knn) estimate of those
probabilities, for a set of simulations drawn at stage (A), whose records constitute the so-called
{\em reference table}, see \cite{biau:etal:2015} or \cite{stoehr:pudlo:cucala:2014}.

\begin{algorithm}[H]
  \caption{ABC model choice algorithm}
  \label{algo:general}
  \begin{itemize} 
  \item[\bf ~(A)] Generate a reference table including $N_\text{ref}$ simulations
    $(m,S(\bx))$ from $\pi(m)\pi(\theta| m)f(\bx|m,\theta)$
  \item[\bf ~(B)] Learn from this set to infer about
    $m$  at $\mathbf{s}^0=S(\mathbf{x}^0)$
  \end{itemize}
\end{algorithm}

Selecting a set of summary statistics $S(x)$ that are informative for model choice is
an important issue. The ABC approximation to the posterior probabilities
$\pi(m|\bx^0)$ will eventually produce a right ordering of the fit of competing models to the
observed data and thus select the right model for a specific class of statistics on large
datasets \citep{marin:pillai:robert:rousseau:2014}. This most recent theoretical ABC model choice results
indeed show that some statistics produce nonsensical decisions and that there exist sufficient
conditions for statistics to produce consistent model prediction, albeit at the cost of an
information loss due to summaries that may be substantial. The toy example comparing MA(1) and MA(2)
models in Appendix and Figure \ref{fig:truePPvssummaries} clearly exhibits this
potential loss in using only the first two autocorrelations as summary statistics.
\cite{barnes:etal:2012} developed an interesting methodology to select the summary statistics, but with
the requirement to aggregate estimation and model pseudo-sufficient statistics for every model under comparison.
That induces a deeply inefficient dimension inflation and can be very time consuming.

\begin{figure}[ht]
  \centering
  \includegraphics[width=0.5\textwidth]{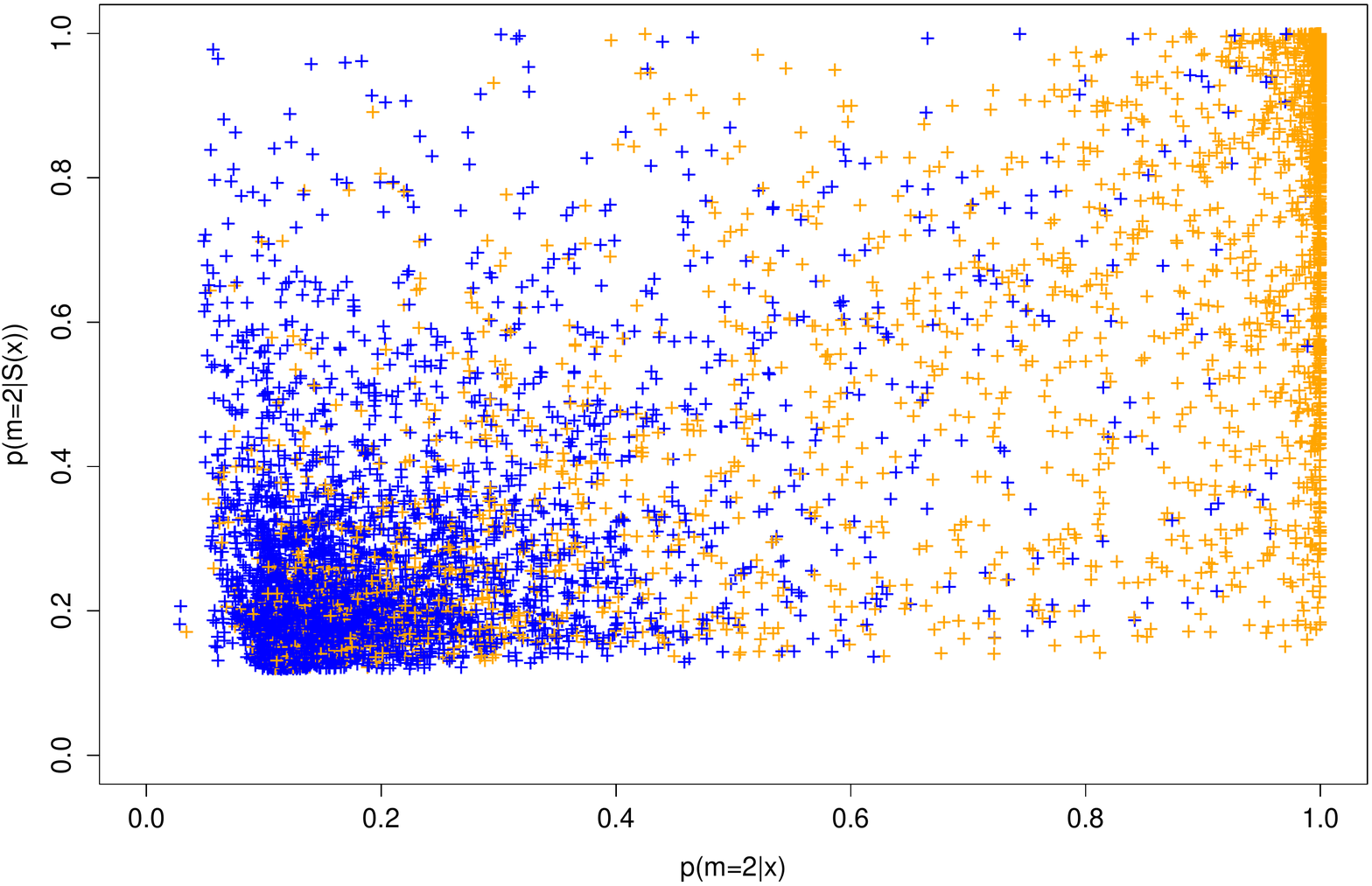}
  
  \vspace{0.5cm} \caption{{\bf Illustration of the discrepancy between posterior probabilities based
    on the whole data and based on a summary.} The aim is to choose between two nested time series models,
    namely moving averages of
    order 1 and 2 (denoted MA(1) and MA(2) respectively; see Appendix for more details). Each point of the 
    plot gives two posterior probabilities of MA(2) for a dataset simulated either
    from the MA(1) (\textit{blue}) or MA(2) model (\textit{orange}), based on the whole data ($x$-axis)
    and on only the first two autocorrelations ($y$-axis).
    \label{fig:truePPvssummaries}
  }
\end{figure}

It may seem tempting to collect the largest possible number of summary statistics to capture more
information from the data. This brings $\pi(m|S(\bx^0))$ closer to $\pi(m|\bx^0)$ but increases the
dimension of $S(\bx)$. ABC algorithms, like \knn~and other local methods suffer from the curse of
dimensionality (see e.g.~Section 2.5 in \cite{hastie:tibshirani:friedman:2009})
so that the estimate of $\pi(m|S(\bx^0))$ based on the simulations is poor when the dimension of
$S(\bx)$ is too large.  Selecting summary statistics correctly and sparsely is therefore paramount, 
as shown by the literature in the recent years. (See \cite{blum:nunes:prangle:sisson:2013} surveying ABC parameter
estimation.) For ABC model choice, two main projection techniques have been considered so far. 
First, \cite{prangle:etal:2014} show that the Bayes factor itself is an acceptable summary (of dimension one) when
comparing two models, but its practical evaluation via a pilot ABC simulation induces a poor approximation of model
evidences \citep{didelot:everitt:johansen:lawson:2011,robert:cornuet:marin:pillai:2011}.  The recourse to a regression
layer like linear discriminant analysis (LDA, \citealp{estoup:etal:2012}) is discussed below and in Appendix Section A.
Other projection techniques have been proposed in the context of parameter estimation: see, e.g.,
\cite{fearnhead:prangle:2012,aeschbacher:beaumont:futschik:2012}.

Given the fundamental difficulty in producing reliable tools for model choice based on summary
statistics \citep{robert:cornuet:marin:pillai:2011}, we now propose to switch to a different approach
based on an adapted classification method.  We recall in the next Section the most important features
of the Random Forest (RF) algorithm.

\subsection{Random Forest methodology}

The Classification And Regression Trees (CART) algorithm at the core of the RF scheme produces a binary tree
that sets allocation rules for entries as labels of the internal nodes and classification or predictions of $Y$ as
values of the tips (terminal nodes). At a given internal node, the binary rule compares a selected covariate $X_j$ with
a bound $t$, with a left-hand branch rising from that vertex defined by $X_j<t$. Predicting the value of $Y$ given the
covariate $X$ implies following a path from the tree root that is driven by applying these binary rules. The outcome of
the prediction is the value found at the final leaf reached at the end of the path: majority rule for classification
and average for regression. To find the best split and the best variable at each node of the tree,
we minimize a criterium: for classification, the Gini index and, for regression, the $L^2$-loss.
In the randomized version of the CART algorithm (see Algorithm A1 in the Appendix),
only a random subset of covariates of size $n_\text{try}$ is considered at each node of the tree.

The RF algorithm \citep{breiman:2001} consists in bagging (which stands for bootstrap aggregating)
randomized CART. It produces $N_\text{tree}$ randomized CART trained on samples
or sub-samples of size $N_\text{boot}$ produced by bootstrapping the original training database. 
Each tree provides a classification or a regression rule that returns a class or a prediction.
Then, for classification we use the majority vote across all trees in the forest, and,
for regression, the response values are averaged.

Three tuning parameters need be calibrated: the number $N_\text{tree}$ of trees in the forest, the number
$n_\text{try}$ of covariates that are sampled at a given node of the randomized CART, and the size $N_\text{boot}$ of
the bootstrap sub-sample. This point will be discussed in Section \ref{sec:prac}.

For classification, a very useful indicator is the \textit{out-of-bag} error 
\citep[][Chapter 15]{hastie:tibshirani:friedman:2009}. 
Without any recourse to a test set, it gives you some idea on how good is your RF classifier.
For each element of the training set, we can define the out-of-bag classifier: the aggregation of votes
over the trees not constructed using this element. The out-of-bag error is the error rate of the out-of-bag
classifier on the training set. The out-of-bag error estimate is as accurate as using a test set of the
same size as the training set.

\subsection{ABC model choice via random forests}

The above-mentioned difficulties in ABC model choice drives us to a paradigm shift in the practice of model choice,
namely to rely on a classification algorithm for model selection, rather than a poorly
estimated vector of $\pi(m|S(\bx^0))$ probabilities. As shown in the example described in Section 3.1, the standard ABC
approximations to posterior probabilities can significantly differ from the true $\pi(m|\bx^0)$. Indeed, our version of
stage (B) in Algorithm~\ref{algo:general} relies on a RF classifier whose goal is to predict the suited model
$\hat{m}(s)$ at each possible value $s$ of the summary statistics $S(\bx)$. The random forest is trained on the 
simulations produced by stage (A) of Algorithm~\ref{algo:general}, which constitute the reference table. Once the model
is selected as $m^\star$, we opt to approximate $\pi(m^*|S(\bx^0))$ by another random forest, obtained from regressing
the probability of error on the (same) covariates, as explained below.

A practical way to evaluate the performance of an ABC model choice algorithm
and to check whether both a given set of summary statistics and a given classifier is to check whether
it provides a better answer than others. The aim is to come near the so-called {\em Bayesian classifier}, which, for the
observed $\bx$, selects the model having the largest posterior probability $\pi(m|\bx)$. It is well known that
the Bayesian classifier (which cannot be derived) minimizes the 0--1 integrated
loss or error \citep{devroye:gyorfi:lugosi:1996}. In the ABC framework, we call the
integrated loss (or risk) the {\em prior error rate}, since it provides an indication of the global quality of a
given classifier $\hat{m}$ on the entire space weighted by the prior. This rate is the expected
value of the misclassification error over the hierarchical prior
\[
  \sum_m \pi(m) \int \mathbf 1{\{\hat{m}(S(\by))\neq m\}}\
  f(\by|\theta,m) \pi(\theta|m) \text{d}\by \text{d}\theta \,.\label{eq:prior.error}
\]
It can be evaluated from simulations $(\theta,m,S(\by))$ drawn as in stage (A) of
Algorithm~\ref{algo:general}, independently of the reference table \citep{stoehr:pudlo:cucala:2014},
or with the out-of-bag error in RF that, as explained above, requires no further simulation.
Both classifiers and sets of summary statistics 
can be compared via this error scale: the pair that minimizes the prior error rate achieves
the best approximation of the ideal Bayesian classifier. In that sense it stands closest to the 
decision we would take were we able to compute the true $\pi(m|\bx)$. 

We seek a classifier in stage (B) of Algorithm~\ref{algo:general} that can handle an
arbitrary number of statistics and extract the maximal information from the reference table obtained
at stage (A). As introduced above, random forest (RF) classifiers \citep{breiman:2001} are perfectly suited
for that purpose. The way we build both a RF classifier given a collection of statistical models and an associated RF
regression function for predicting the allocation error is to start from a simulated ABC {\em reference table}
made of a set of simulation records made of model indices and summary statistics for the
associated simulated data. This table then serves as training database for a RF that forecasts model index based on
the summary statistics. The resulting algorithm, presented in Algorithm \ref{algo:ABC-RF} and called ABC-RF,
is implemented in the R package {\sf abcrf} associated with this paper.

\begin{algorithm}
   \caption{\sffamily ABC-RF}
  \label{algo:ABC-RF}
  \begin{algorithmic} 
  \item[\bf ~(A)] Generate a reference table including $N_\text{ref}$ simulation $(m,S(\bx))$
  from $\pi(m)\pi(\theta| m)f(\bx|m,\theta)$
  \item[\bf ~(B)] Construct $N_\text{tree}$ randomized CART which predict $m$ using $S(\bx)$
    \FOR{$b=1$ \TO $N_\text{tree}$}
    \STATE \textbf{draw} a bootstrap (sub-)sample of size $N_\text{boot}$ from the reference table
    \STATE \textbf{grow} a randomized CART $T_b$ (Algorithm A1 in the Appendix) 
    \ENDFOR
  \item[\bf ~(C)] Determine the predicted indexes for $S(\bx^0)$ and the trees $\{T_b;b=1,\ldots,N_\text{tree}\}$
  \item[\bf ~(D)] Affect $S(\bx^0)$ according to a majority vote among the predicted indexes
  \end{algorithmic}
\end{algorithm}

The justification for choosing RF to conduct an ABC model selection is that, both formally
\citep{biau:2012,scornet:biau:vert:2014} and experimentally \citep[][Chapter
5]{hastie:tibshirani:friedman:2009}, RF classification was shown to be mostly insensitive both to
strong correlations between predictors (here the summary statistics) and to the presence of noisy
variables, even in relatively large numbers, a characteristic that \knn\ classifiers
miss. 

This type of robustness justifies adopting an RF strategy to learn from an ABC reference table for
Bayesian model selection. Within an arbitrary (and arbitrarily large) collection of summary statistics, some may exhibit
strong correlations and others may be uninformative about the model index, with no terminal consequences on
the RF performances. For model selection, RF thus competes with both local classifiers
commonly implemented within ABC: It provides a more non-parametric modelling than local logistic regression
\citep{beaumont:2008}, which is implemented in the DIYABC software \citep{cornuet:etal:2014} which is extremely
costly --- see, e.g., \cite{estoup:etal:2012} which reduces the dimension using linear discriminant projection before
resorting to local logistic regression. This software also includes a standard
\knn\ selection procedure which suffers from the curse of
dimensionality and thus forces selection among statistics.

\subsection{Approximating the posterior probability of the selected model}

The outcome of RF computation applied to a given target dataset is a classification vote for each model
which represents the number of times a model is selected in a forest of n trees. The model with the highest
classification vote corresponds to the model best suited to the target dataset. It is worth stressing here that there
is no direct connection between the frequencies of the model allocations of the data among the tree classifiers
(i.e. the classification vote) and the posterior probabilities of the competing models.
Machine learning classifiers hence miss a distinct advantage of posterior probabilities, namely that the
latter evaluate a confidence degree in the selected (MAP) model.  An alternative to those
probabilities is the prior error rate. Aside from its use to select the best classifier and set of summary statistics,
this indicator remains, however, poorly relevant since the only point of importance in the
data space is the observed dataset $S(\bx^0)$.

A first step addressing this issue is to obtain error rates conditional on the data as in
\cite{stoehr:pudlo:cucala:2014}. However, the statistical methodology considered therein 
suffers from the curse of dimensionality and we here consider a different approach to precisely estimate this
error. We recall \citep{robert:2001} that the posterior probability of a model is the natural Bayesian uncertainty
quantification since it is the complement of the posterior error associated with the loss
$\mathbb{I}(\hat m(S(\bx^0))\ne m)$.
While the proposal of \cite{stoehr:pudlo:cucala:2014} for estimating the conditional error rate
induced a classifier given $S=S(\bx^0)$
\begin{equation}\label{rouge}
\prob\big(\hat{m}(S(Y)) \neq m \big| S(Y)=S(\bx^0)\big) \,,
\end{equation}
involves non-parametric kernel regression, we suggest to rely instead on a RF regression to undertake this
estimation. The curse of dimensionality is then felt much less acutely, given that random forests can accommodate large
dimensional summary statistics. Furthermore, the inclusion of many summary statistics does not induce a reduced
efficiency in the RF predictors, while practically compensating for insufficiency.

Before describing in more details the implementation of this concept, we stress that the perspective of
\cite{stoehr:pudlo:cucala:2014} leads to effectively estimate the posterior probability that the true model is the MAP,
thus providing us with a non-parametric estimation of this quantity, an alternative to the classical ABC solutions we
found we could not trust. Indeed, the posterior expectation \eqref{rouge} satisfies
\newcommand\sobs{\mathbf{s}^0}
\begin{align*}
\mathbb{E}[\mathbb{I}(\hat{m}(\sobs)\ne m)|S(\bx^0)]
&=\sum_{i=1}^k \mathbb{E}[\mathbb{I}(\hat{m}(S(\bx^0))\ne m=i)|S(\bx^0)]\\
&=\sum_{i=1}^k \mathbb{P}[m=i)|S(\bx^0)]\times\mathbb{I}(\hat{m}(S(\bx^0))\ne i)\\
&=\mathbb{P}[m\ne \hat{m}(S(\bx^0))|S(\bx^0)]\\
&=1-\mathbb{P}[m = \hat{m}(S(\bx^0))|S(\bx^0)]\,.
\end{align*}
It therefore provides the complement of the posterior probability that the true model is the selected model.

To produce our estimate of the posterior probability $\mathbb{P}[m = \hat{m}(S(\bx^0))|S(\bx^0)]$, we
proceed as follows:
\begin{enumerate}
\item we compute the value of $\mathbb{I}(\hat{m}(s)\ne m)$ for the trained random forest $\hat{m}$ and 
for all terms in the ABC reference table; to avoid overfitting, we use the out-of-bag classifiers;
\item we train a RF regression estimating the variate $\mathbb{I}(\hat{m}(s)\ne m)$ as a function of
the same set of summary statistics, based on the same reference table. This second RF can be represented as 
a function $\varrho(s)$ that constitutes a machine learning estimate of $\mathbb{P}[m\ne \hat{m}(s)|s]$;
\item we apply this RF function to the actual observations summarized as $S(\bx^0)$ and return
$1-\varrho(S(\bx^0))$  as our estimate of $\mathbb{P}[ m = \hat{m}(S(\bx^0))|S(\bx^0)]$.
\end{enumerate}
This corresponds to the representation of Algorithm \ref{algo:posterior}
which is implemented in the R package {\sf abcrf} associated with this paper.

\begin{algorithm}
\caption{Estimating the posterior probability of the selected model}
\label{algo:posterior}
\begin{itemize}
\item[(a)] Use the RF produce by Algorithm \ref{algo:ABC-RF} to compute the out-of-bag classifiers
of all terms in the reference table and deduce the associated binary model prediction error
\item[(b)] Use the reference table to build a RF regression function $\varrho(s)$  
regressing the model prediction error on the summary statistics
\item[(c)] Return the value of $1-\varrho(S(\bx^0))$ as the RF regression estimate of
$\mathbb{P}[ m = \hat{m}(S(\bx^0))|S(\bx^0)]$
\end{itemize}
\end{algorithm}

\section{Results: illustrations of the ABC-RF methodology}

To illustrate the power of the ABC-RF methodology, we now report several controlled experiments as well as two
genuine population genetic examples. 

\subsection{Insights from controlled experiments}

The appendix details controlled experiments on a toy problem, comparing MA(1) and MA(2) time-series models,
and two controlled synthetic examples from population genetics, based on Single Nucleotide Polymorphism
(SNP) and microsatellite data. The toy example is particularly revealing with regard to
the discrepancy between the posterior probability
of a model and the version conditioning on the summary statistics $S(\bx^0)$. 
Figure \ref{fig:truePPvssummaries} shows how far from the diagonal are realizations of the
pairs $(\pi(m|\bx^0),\pi(m|S(\bx^0)))$, even though the autocorrelation statistic is quite informative
\citep{marin:pudlo:robert:ryder:2012}. Note in particular the vertical accumulation of points near
$\mathbb{P}(m=2|\bx^0)=1$.  Table \ref{tab:MAMA} in the appendix demonstrates the further gap in predictive power for the full Bayes
solution with a true error rate of 12\%~versus the best solution (RF) based on the summaries barely
achieving a 16\%~error rate.

For both controlled genetics experiments in the appendix, the computation of the true posterior probabilities
of the three models is impossible. The predictive performances of the competing classifiers can
nonetheless be compared on a test sample. Results, summarized in Tables \ref{tab:snp} and \ref{tab:microsat} in the appendix,
legitimize the use of RF, as this method achieves the most efficient classification in all genetic
experiments. Note that that the prior error rate of any classifier is always bounded from below 
by the error rate associated with the (ideal) Bayesian classifier. Therefore, a mere gain of a few percents may well  
constitute an important improvement when the prior error rate is low. As an aside, we also stress that, since
the prior error rate is an expectation over the entire sampling space, the reported gain may occult
much better performances over some areas of this space.

Figure \ref{fig:MAMA.post} in the appendix displays differences between the true posterior probability of 
the model selected by Algorithm \ref{algo:ABC-RF} and its approximation with Algorithm \ref{algo:posterior}.

\subsection{Microsatellite dataset: retracing the invasion routes of the Harlequin ladybird}

The original challenge was to conduct inference about the introduction pathway of the invasive
Harlequin ladybird \textit{(Harmonia axyridis}) for the first recorded outbreak of this species in
eastern North America.  The dataset, first analyzed in \cite{lombaert:etal:2011} and
\cite{estoup:etal:2012} via ABC, includes samples from three natural and two biocontrol populations
genotyped at 18 microsatellite markers. The model selection requires the formalization and
comparison of 10 complex competing scenarios corresponding to various possible routes of
introduction (see appendix for details and analysis 1 in \cite{lombaert:etal:2011}).  We now compare our
results from the ABC-RF algorithm with other classification methods for three sizes of the reference
table and with the original solutions by \cite{lombaert:etal:2011} and \cite{estoup:etal:2012}.  
We included all summary statistics computed by the DIYABC software for microsatellite
markers \citep{cornuet:etal:2014}, namely 130 statistics, complemented by the nine LDA axes as
additional summary statistics (see appendix Section G).

In this example, discriminating among models based on the observation of summary statistics is
difficult.  The overlapping groups of Figure \ref{fig:cox_lda} in the appendix reflect that difficulty, the source of which is
the relatively low information carried by the 18 autosomal microsatellite loci considered here.
Prior error rates of learning methods on the whole reference table are given in
Table \ref{tab:asian}. As expected in such a high dimension settings \cite[][Section
2.5]{hastie:tibshirani:friedman:2009}, \knn\ classifiers behind the standard ABC methods are all
defeated by RF for the three sizes of the reference table, even when \knn\ is trained on the much smaller set of covariates composed of the nine
LDA axes.  The classifier and set of summary statistics showing the lowest prior error rate
is RF trained on the 130 summaries and the nine LDA axes. 

Figure \ref{fig:cox_viss} in the appendix shows that RFs are able to automatically determine the (most) relevant statistics for model comparison,
including in particular some crude estimates of admixture rate defined in \cite{choisy:etal:2004},
some of them not selected by the experts in \cite{lombaert:etal:2011}. We stress here that the level
of information of the summary statistics displayed in Figure \ref{fig:cox_viss} in the appendix is relevant for model choice but not
for parameter estimation issues. In other words, the set of best summaries found with ABC-RF should
not be considered as an optimal set for further parameter estimations under a given model with
standard ABC techniques \citep{beaumont:zhang:balding:2002}.

\begin{table}
\small
\caption{\textbf{Harlequin ladybird data}: estimated prior error rates for various classification
methods and   sizes of the reference table.\label{tab:asian}}
\begin{center}
\begin{tabular}{crrr}
\hline 
\textbf{Classification method } & 
\multicolumn{3}{c}{\textbf{Prior error rates} (\%)} \\ 
  \textbf{trained on} & $N_\text{ref}=10,000$ & $N_\text{ref}=20,000$ & $N_\text{ref}=50,000$
\\
\hline
Linear discriminant analysis (LDA)       & $39.91$ & $39.30$ & $39.04$ \\
Standard ABC (\knn) on DIYABC summaries  & $57.46$ & $53.76$ & $51.03$ \\
Standard ABC (\knn) on LDA axes          & $39.18$ & $38.46$ & $37.91$ \\              
Local logistic regression on LDA axes    & $41.04$ & $37.08$ & $36.05$ \\
RF on DIYABC summaries                   & $40.18$ & $38.94$ & $37.63$ \\
RF on DIYABC summaries and LDA axes      & $36.86$ & $35.62$ & $34.44$ \\
\hline
\end{tabular}
\end{center}
\begin{flushleft} 
\footnotesize
  Note - Performances of classifiers used in stage (B) of Algorithm~\ref{algo:general}. A set of $10,000$
  prior simulations was used to calibrate the number of neighbors $k$ in both standard ABC and local
  logistic regression. Prior error rates are estimated as average misclassification errors on an independent
  set of $10,000$ prior simulations, constant over methods and sizes of the reference tables. 
  $N_\text{ref}$ corresponds to the number of simulations included in the reference table. 
\end{flushleft}
\end{table}

The evolutionary scenario selected by our RF strategy agrees with the earlier conclusion of
\cite{lombaert:etal:2011}, based on approximations of posterior probabilities with local logistic
regression solely on the LDA axes \textit{i.e.}, the same scenario displays the highest ABC
posterior probability and the largest number of selection among the decisions taken by the
aggregated trees of RF. Using Algorithm \ref{algo:posterior}, we got an estimate of the posterior
probability of the selected scenario equal to $0.4624$. This estimate is significantly lower than
the one of about $0.6$ given in \cite{lombaert:etal:2011} based on a local logistic regression method.
This new value is more credible because: it is based on all the summary statistics and, on a method
adapted to such an high dimensional context and less sensible to calibration issues. Moreover, 
this small posterior probability corresponds better to the intuition of the experimenters and
indicates that new experiments are necessary to give a more reliable answer.

\subsection{SNP dataset: inference about Human population history}

Because the ABC-RF algorithm performs well with a substantially lower number of simulations compared to standard
ABC methods, it is expected to be of particular interest for the statistical processing of massive
single nucleotide polymorphism (SNP) datasets, whose production is on the increase in the field of
population genetics. We analyze here a dataset including 50,000 SNP markers genotyped in four Human
populations \citep{genome:project:2012}. The four populations include Yoruba (Africa), Han (East
Asia), British (Europe) and American individuals of African ancestry, respectively. Our intention is
not to bring new insights into Human population history, which has been and is still studied in
greater details in research using genetic data, but to illustrate the potential of ABC-RF in this
context. We compared six scenarios (i.e. models) of evolution of the four Human populations which
differ from each other by one ancient and one recent historical events: (i) a single out-of-Africa
colonization event giving an ancestral out-of-Africa population which secondarily split into one
European and one East Asian population lineages, versus two independent out-of-Africa colonization
events, one giving the European lineage and the other one giving the East Asian lineage; (ii) the
possibility of a recent genetic admixture of Americans of African origin with their African
ancestors and individuals of European or East Asia origins. The SNP dataset and the compared
scenarios are further detailed in the appendix. We used all the summary statistics provided by DIYABC for
SNP markers \citep{cornuet:etal:2014}, namely 112 statistics in this setting complemented by the five
LDA axes as additional statistics.

\begin{figure}[ht!]
\centering
  \includegraphics[width=0.5\textwidth]{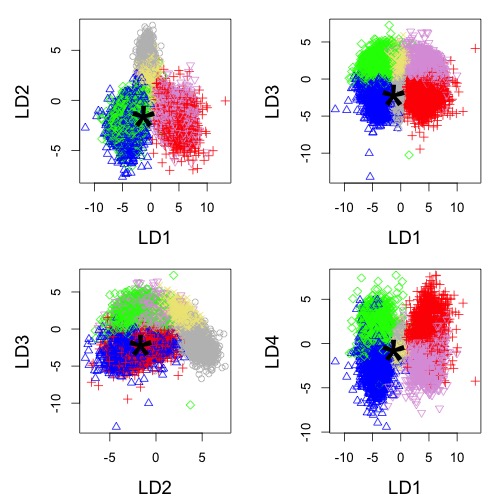}
  
  \vspace{0.5cm}\caption{{\bf Human SNP data}: projection of the reference table
  on the first four LDA axes. Colors correspond to model indices.
  The location of the additional datasets is indicated by a {\em large black star}. \label{fig:human_lda}}
\end{figure}

To discriminate between the six scenarios of Figure \ref{fig:outofAf} in the appendix,
RF and others classifiers have been trained on three reference tables of different sizes.
The estimated prior error rates are reported in Table~\ref{tab:human}. Unlike the previous example,
the information carried here by the $50,000$ SNP markers is much higher, because it induces better
separated simulations on the LDA axes (Figure \ref{fig:human_lda}), and much lower prior error rates
(Table~\ref{tab:human}). RF using both the initial summaries and the LDA axes provides the best results.

\begin{table}
\small
\caption{{\bf Human SNP data}: estimated prior error rates for classification methods
  and three sizes of reference table.
\label{tab:human}}
\begin{center}
\begin{tabular}{crrr}
\hline 
\textbf{Classification method } & 
\multicolumn{3}{c}{\textbf{Prior error rates} (\%) } \\ 
  \textbf{trained on} & $N_\text{ref}=10,000$ & $N_\text{ref}=20,000$ & $N_\text{ref}=50,000$ \\
\hline 
Linear discriminant analysis (LDA)          & $9.91$  & $9.97$  & $10.03$ \\
Standard ABC (\knn) using DYIABC summaries  & $23.18$ & $20.55$ & $17.76$ \\
Standard ABC (\knn) using only LDA axes     & $6.29$  & $5.76$  & $5.70$ \\ 
Local logistic regression on LDA axes       & $6.85$  & $6.42$  & $6.07$ \\
RF using DYIABC initial summaries           & $8.84$  & $7.32$  & $6.34$ \\
RF using both DIYABC summaries and LDA axes & $5.01$  & $4.66$  & $4.18$ \\
\hline
\end{tabular}
\end{center}
\begin{flushleft} 
\footnotesize Note - Same comments as in Table \ref{tab:asian}.
\end{flushleft}
\end{table}

ABC-RF algorithm selects Scenario 2 as the forecasted scenario on the Human dataset, an answer which is not
visually obvious on the LDA projections of Figure \ref{fig:human_lda}, Scenario 2 corresponds to the blue color.
But considering previous population genetics studies in the field, it is not surprising that this scenario, which includes a
single out-of-Africa colonization event giving an ancestral out-of-Africa population with a
secondarily split into one European and one East Asian population lineage and a recent genetic
admixture of Americans of African origin with their African ancestors and European individuals, was
selected. Using Algorithm \ref{algo:posterior}, we got an estimate of the posterior probability 
of scenario 2 equal to $0.998$ which is as expected very high.

Computation time is a particularly important issue in the present example.  Simulating the $10,000$
SNP datasets used to train the classification methods requires seven hours on a computer with 32
processors (Intel Xeon(R) CPU 2GHz).  In that context, it is worth stressing that RF
trained on the DIYABC summaries and the LDA axes of a $10,000$ reference table has a
smaller prior error rate than all other classifiers, even when they are trained on a $50,000$
reference table. In practice, standard ABC treatments for model choice are based on reference
tables of substantially larger sizes (i.e. $10^5$ to $10^6$ simulations per scenario 
\citep{estoup:etal:2012,bertorelle:etal:2010}).
For the above setting in which six scenarios are
compared, standard ABC treatments would hence request a minimum computation time of 17 days (using the
same computation resources). According to the comparative tests that we carried out on various
example datasets, we found that RF globally allowed a minimum computation speed gain around a factor
of 50 in comparison to standard ABC treatments.

\subsection{Practical recommendations regarding the implementation of the algorithms}
\label{sec:prac}

We develop here several points, formalized as questions, which should help users seeking to
apply our methodology on their dataset for statistical model choice.

\subsubsection*{Are my models and/or associated priors compatible with the observed dataset?}

This question is of prime interest and applies to any type of ABC treatment, including both standard ABC treatments and
treatments based on ABC random forests. Basically, if none of the proposed model - prior combinations produces some
simulated datasets in a reasonable vicinity of the observed dataset, it is a signal of incompatibility and we consider
it is then useless to attempt model choice inference. In such situations, we strongly advise reformulating the compared
models and/or the associated prior distributions in order to achieve some compatibility in the above sense. We propose
here a visual way to address this issue, namely through the simultaneous projection of the simulated reference table
datasets and of the observed dataset on the first LDA axes, such a graphical assessment can be achieved using the R
package {\sf abcrf} associated with this paper. In the LDA projection, the observed dataset need be located reasonably
within the clouds of simulated datasets (see Figure 2 as an illustration).  Note that visual representations of a
similar type (although based on PCA) as well as computation for each summary statistics and for each model of the
probabilities of the observed values in the prior distributions have been proposed by \cite{cornuet:ravigne:estoup:2010}
and are already automatically provided by the DIYABC software.

\subsubsection*{Did I simulate enough datasets for my reference table?}

A rule of thumb is to simulate between 5,000 and 10,000 datasets per model among those compared. For instance, in the
example dealing with Human population history (Section 3.3) we have simulated a total of 50,000 datasets from six models
(i.e., about 8,300 datasets per model). To evaluate whether or not this number is sufficient for random forest analysis,
we recommend to compute global prior error rates from both the entire reference table and a subset of the reference
table (for instance from a subset of 40,000 simulated datasets if the reference table includes a total of 50,000
simulated datasets). If the prior error rate value obtained from the subset of the reference
table is similar, or only lightly higher, than the value obtained from the entire reference table, 
one can consider that the reference table
contains enough simulated datasets. If a substantial difference is observed between both values, then we recommend an
increase in the number of datasets in the reference table. For instance, in the Human population history example we
obtained prior error rate values of 4.22\% and 4.18\% when computed from a subset of 40,000
simulated datasets and the entire 50,000 datasets of the reference table, respectively.
In this case, the hardship of producing more simulated dataset in the reference table seems negligible.

\subsubsection*{Did my forest grow enough trees?}

\begin{figure}[ht!]
\centering
  \includegraphics[width=0.5\textwidth]{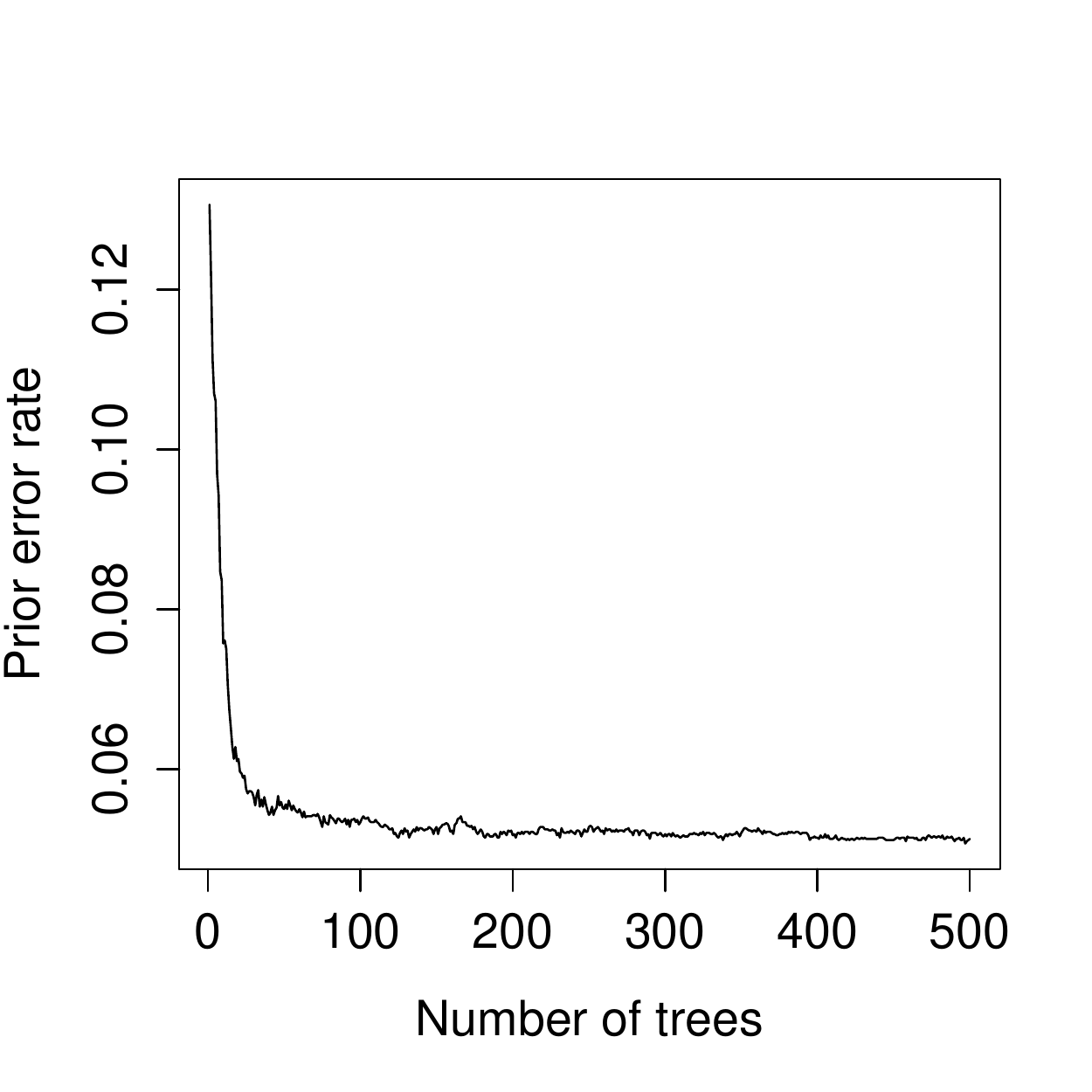}
  
  \vspace{0.5cm}\caption{{\bf Human SNP data}: evolution of the ABC-RF prior error rate
  when $N_\text{ref}=50,000$ with respect to the number of trees in the forest. \label{fig:human_prior}}
\end{figure}

According to our experience, a forest made of 500 trees usually constitutes \citep{breiman:2001} an interesting
trade-off between computation efficiency and statistical precision. To evaluate whether or not this number is
sufficient, we recommend to plot the estimated values of the prior error rate and/or the posterior probability of the
best model as a function of the number of trees in the forest. The shapes of the curves provide a visual diagnostic of
whether such key quantities stabilize when the number of trees tends to 500. We provide illustrations of such visual
representations in the case of the example dealing with Human population history in Figure \ref{fig:human_prior}. 

\subsubsection*{How do I set $N_\text{boot}$ and $N_\text{try}$? for classification and regression}

For a reference table with up to 100,000 datasets and 250 summary statistics, we recommend keeping the entire reference
table, that is, $N_\text{boot}=N$ when building the trees. For larger reference tables, the value of $N_\text{boot}$ can
be calibrated against the prior error rate, starting with a value of $N_\text{boot}=50,000$ and doubling it until the
estimated prior error rate is stabilized. For the number $n_\text{try}$ of summary statistics sampled at each of the
nodes, we see no reason to modify the default number of covariates $n_\text{try}$ is chosen as $\sqrt{d}$ for classification
and $d/3$ for regression when $d$ is the total number of predictors. 
Finally, when the number of summary statistics is lower than 15, one might reduce $N_\text{boot}$ to $N/10$.

\section{Discussion}\label{sec:on}

The present paper is purposely focused on selecting a statistical model, which can be rephrased as a classification
problem trained on ABC simulations. We defend here the paradigm shift of assessing the best fitting model via a random
forest classification and in evaluating our confidence in the selected model by a secondary random forest procedure,
resulting in a different approach to precisely estimate the posterior probability of the selected model. 
We further provide a calibrating principle for this
approach, in that the prior error rate provides a rational way to select the classifier and the set of summary
statistics which leads to results closer to a true Bayesian analysis. 

Compared with past ABC implementations, ABC-RF offers improvements at least at four levels: (i) on
all experiments we studied, it has a lower prior error rate; (ii) it is robust to the size and
choice of summary statistics, as RF can handle many superfluous statistics with no impact on the
performance rates (which mostly depend on the intrinsic dimension of the classification problem
\citep{biau:2012,scornet:biau:vert:2014}, a characteristic confirmed by our results); (iii) the
computing effort is considerably reduced as RF requires a much smaller reference table compared with
alternatives (\textit{i.e.}, a few thousands versus hundred thousands to billions of simulations);
and (iv) the method is associated with an embedded and free error evaluation which assesses the
reliability of ABC-RF analysis.  As a consequence, ABC-RF allows for a more robust handling of the
degree of uncertainty in the choice between models, possibly in contrast with earlier and
over-optimistic assessments.

Due to a massive gain in computing and simulation efforts, ABC-RF will extend the range and
complexity of datasets (e.g. number of markers in population genetics) and models handled by
ABC. In particular, we believe that ABC-RF will be of considerable interest for the statistical processing
of massive SNP datasets whose production rapidly increases within the field of population genetics
for both model and non-model organisms.
Once a given model has been chosen and confidence evaluated by ABC-RF, it becomes possible to
estimate parameter distribution under this (single) model using standard ABC techniques
\citep{beaumont:zhang:balding:2002} or alternative methods such as those proposed by
\cite{excoffier:etal:2013}.

\section{Acknowledgments}
The use of random forests was suggested to JMM and CPR by Bin Yu during a visit at CREST, Paris.  We are grateful
to G. Biau for his help about the asymptotics of random forests.  Some parts of the research were conducted at
BIRS, Banff, Canada, and the authors (PP and CPR) took advantage of this congenial research environment.
This research was partly funded by the ERA-Net BiodivERsA2013-48 (EXOTIC), with the national funders FRB, ANR, MEDDE, BELSPO,
PT-DLR and DFG, part of the 2012-2013 BiodivERsA call for research proposals.

\newpage
\appendix

\centerline{\Large \bf Appendix}

\section{Classification and regression methods}

Classification methods aim at forecasting a variable $Y$ that takes values in a finite set, e.g. $\{1,\ldots,
M\}$, based on a predicting vector of covariates $X=(X_1,\ldots, X_d)$ of dimension $d$. They are fitted with
a training database $(x^i, y^i)$ of independent replicates of the pair $(X,Y)$. We exploit such classifiers in
ABC model choice by predicting a model index ($Y$) from the observation of summary statistics on the data ($X$). The
classifiers are trained with numerous simulations from the hierarchical Bayes model that constitute the ABC
reference table.  For a more detailed entry on classification, we refer the reader to the entry
\cite{hastie:tibshirani:friedman:2009} and to the more theoretical \cite{devroye:gyorfi:lugosi:1996}.

On the other hand, regression methods forecast a continuous variable $Y$ based on the vector
$X$. When trained on a database of independent replicates of a pair $(X, Y)$ with the $L^2$-loss,
regression methods approximate the conditional expected value $\esp(Y|X)$. A random forest for
regression is used to obtain approximations of the posterior probability of the selected model.

\subsection{Standard classifiers} 

Discriminant analysis covers a first family of classifiers including \textit{linear discriminant analysis}
(LDA) and \textit{na{\"\i}ve Bayes}. Those classifiers rely on a full likelihood function corresponding to the
joint distribution of $(X,Y)$, specified by the marginal probabilities of $Y$ and the conditional density
$f(x|y)$ of $X$ given $Y=y$. Classification follows by ordering the probabilities $\text{Pr}(Y=y|X=x)$. For
instance, linear discriminant analysis assumes that each conditional distribution of $X$ is a multivariate
Gaussian distribution with unknown mean and covariance matrix, when the covariance matrix is assumed to be
constant across classes. These parameters are fitted on a training database by maximum likelihood; see
e.g. Chapter 4 of \cite{hastie:tibshirani:friedman:2009}.  This classification method is quite popular as it
provides a linear projection of the covariates on a space of dimension $M-1$, called the LDA axes, which
separate classes as much as possible.  Similarly, \textit{na{\"\i}ve Bayes} assumes that each density
$f(x|y)$, $y=1,\ldots, M$, is a product of marginal densities. Despite this rather strong assumption of conditional
independence of the components of $X$, \textit{na{\"\i}ve Bayes} often produces good classification
results. Typically we assume that the marginals are univariate Gaussians and fit those by maximum
likelihood estimation.

\textit{Logistic and multinomial regressions} use a conditional likelihood based on a modeling of
$\text{Pr}(Y=y|X=x)$, as special cases of a generalized linear model. Modulo a logit transform
$\phi(p)=\log\{p/(1-p)\}$, this model assume linear dependency in the covariates; see e.g. Chapter~4 in
\cite{hastie:tibshirani:friedman:2009}. Logistic regression results rarely differ from LDA estimates since the
decision boundaries are also linear. The sole difference stands with the procedure used to fit the classifiers.

\subsection{Local classification methods} 

\textit{$k$-nearest neighbor} (\knn) classifiers require no model fitting but mere computations on the
training database. More precisely, it builds upon a distance on the feature space, $\mathcal{X}\ni X$.
In order to make a classification when $X=x$, \knn\ derives the $k$ training points that are the closest in distance
to $x$ and classifies this new datapoint $x$ according to a majority vote among the classes of the $k$ neighbors. The accuracy
of \knn\ heavily depends on the tuning of $k$, which should be calibrated, as explained below.

\textit{Local logistic (or multinomial)} regression adds a linear regression layer to these procedures and
dates back to \cite{cleveland:1979}. In order to make a decision at $X=x$, given the $k$ nearest neighbors
in the feature space, one weights them by a smoothing kernel (e.g., the Epanechnikov kernel) and 
a multinomial classifier is then fitted on this weighted sub-sample of the training database. More details on this procedure
can be found in \cite{estoup:etal:2012}. Likewise, the accuracy of the classifier depends on the calibration of $k$.

\subsection{Randomized CART}

The Random Forest (RF) algorithm aggregates randomized Classification And Regression Trees (CART),
which can be trained for both classification and regression issues.  The randomized CART algorithm
used to create the trees in the
forest recursively infers the internal and terminal labels of each tree $i$ from the root on a
training database $(x^i, y^i)$ as follows. Given a tree built until a node $v$, daughter nodes $v_1$
and $v_2$ are determined by partitioning the data remaining at $v$ in a way highly correlated with
the outcome $Y$. Practically, this means minimizing an empirical divergence criterion
towards selecting the best covariate $X_j$ among a random subset of size $n_\text{try}$ and the best
threshold $t$. The covariate $X_j$ and the threshold $t$ is determined by minimizing 
$N(v_1) Q(v_1)+N(v_2) Q(v_2)$, where $N(v_i)$ is the number of records from the training database that fall into node $v_i$ and
$Q(v_i)$ is the empirical criterium at node $v_i$. For regression, we used the $L^2$-loss. For classification,
we privileged the Gini index, defined as
\[
Q(v_i)=\sum_{y=1}^M \hat p(v_i,y)\left\{1-\hat p(v_,y)\right\}\,,
\]
where $\hat p(v_i,y)$ is the relative frequency of $y$ among the part of the
learning database that falls at node $v_i$. Other criteria are possible,
see Chapter 9 in \cite{hastie:tibshirani:friedman:2009}.

For regression, the recursive algorithm stops when all terminal nodes $v$ correspond to at most
five records of the training database, and the label of the tip $v$ is the average of $y$ over these
five records. For classification, it stops when the terminal nodes are homogeneous, {i.e.},
$Q(v)=\sum_{y=1}^M \hat
 p(v, y)\{1-\hat p(v, y)\} = 0$ and the label of the tip $v$ is the only
value of $y$ for which $\hat p(v, y)=1$. This leads to Algorithm S1.

\vspace{0.25cm}\begin{algohere}
  \centerline{\sffamily \normalsize Algorithm S1: {\bf Randomized CART}}
  \begin{algorithmic}\sffamily
    \STATE \textbf{start} the tree with a single root
    \REPEAT
      \STATE \textbf{pick} a non-homogeneous tip $v$ such that $Q(v)>0$ (classification) or 
      $N(v)>5$ (regression)
      \STATE \textbf{attach} to $v$ two daughter nodes $v_1$ and $v_2$
      \STATE \textbf{draw} a random subset of covariates of size $n_\text{try}$
      \FORALL{covariates $X_j$ in the random subset}
        \STATE \textbf{find} the threshold $t_j$ in the rule $X_j<t_j$ that minimizes $N(v_1)Q(v_1)+N(v_2)Q(v_2)$ 
      \ENDFOR
      \STATE \textbf{find} the rule $X_j<t_j$ that minimizes $N(v_1)Q(v_1)+N(v_2)Q(v_2)$ in $j$
      \STATE \AND \textbf{set} this best rule to node $v$
    \UNTIL{all tips $v$ are homogeneous (classification) or correspond to 5 records (regression)}
    \STATE \textbf{set} the labels of all tips
  \end{algorithmic}
\end{algohere}

\subsection{Calibration of the tuning parameters} 

Many machine learning algorithms involve tuning parameters that need to be determined carefully in order to
obtain good results (in terms of what is called the prior error rate in the main text). Usually, the predictive
performances (averaged over the prior in our context) of classifiers are evaluated on new data (validation
procedures) or fake new data (cross-validation procedures); see e.g. Chapter 7 of
\cite{hastie:tibshirani:friedman:2009}. This is the standard way to compare the performances of various
possible values of the tuning parameters and thus calibrate these parameters. For instance, the value of $k$ for both \knn\ and local logistic regression
need to be calibrated. \knn\ performances heavily depend on the value of $k$ 
as illustrated on Figure \ref{fig:elbow}. The plots in this Figure display an empirical evaluation of the prior error
rates of the classifiers against different values of their tuning parameter with a validation sample made of a
fresh set of $10^4$ simulations from the hierarchical Bayesian model. Because of the moderate Monte Carlo
noise within the empirical error, we first smooth out the curve before determining the calibration of the
algorithms. Figure \ref{fig:elbow} displays this derivation for the ABC analysis of the Harlequin ladybird
data with machine learning tools. 

The validation procedure described above requires new simulations from the hierarchical Bayesian model, which we
can always produce because of the very nature of ABC. But such simulations might be computationally intensive when
analyzing large datasets or complex models. Moreover, calibrating local logistic regression may prove computationally
unfeasible since for each dataset of the validation sample (the second reference table), the procedure involves searching for
nearest neighbors in the (first) reference table, then fitting a weighted logistic regression on those neighbors.

\begin{figurehere}
  \begin{center}
  \includegraphics[width=0.4\textwidth]{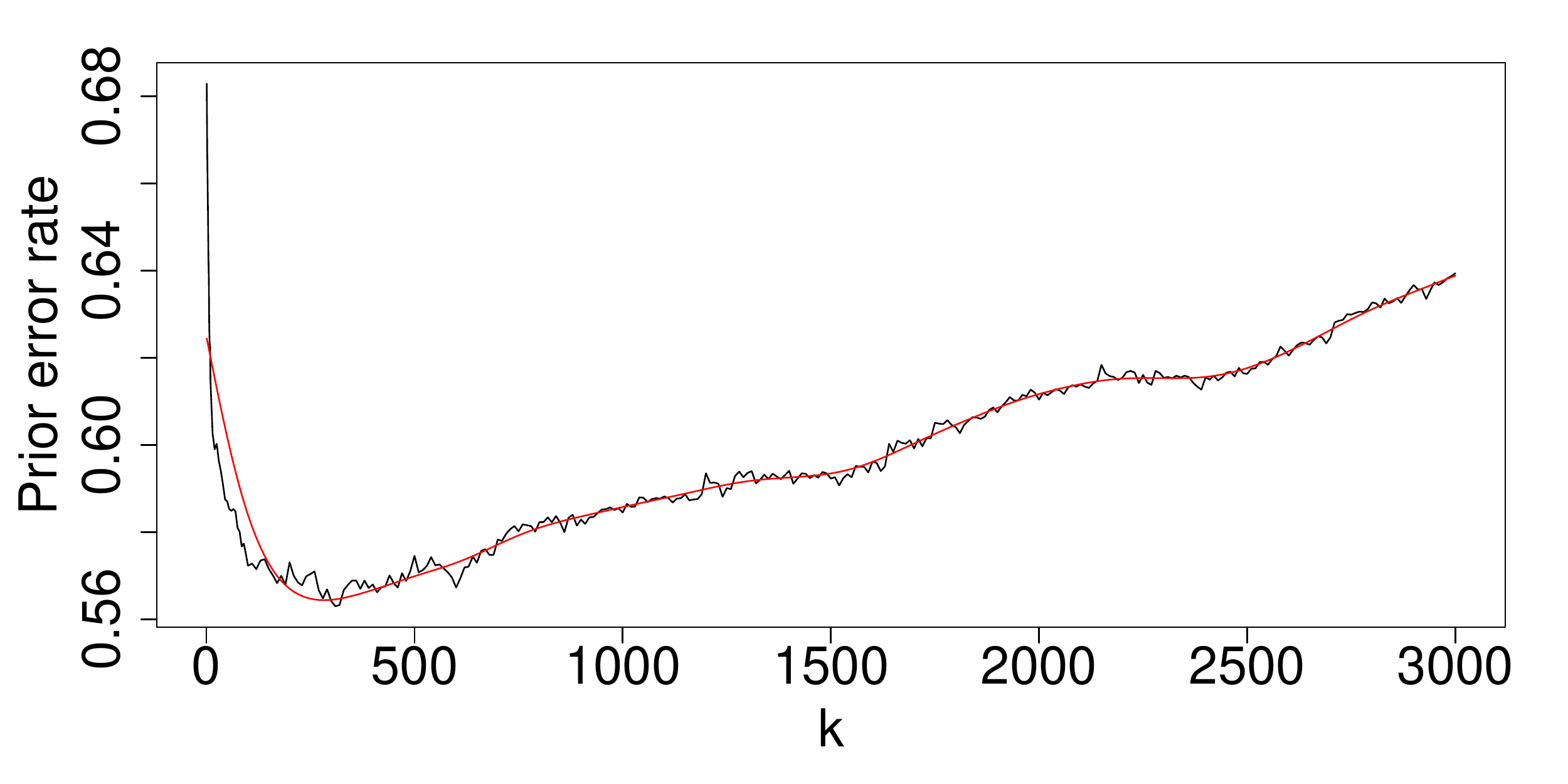} \\
  \includegraphics[width=0.4\textwidth]{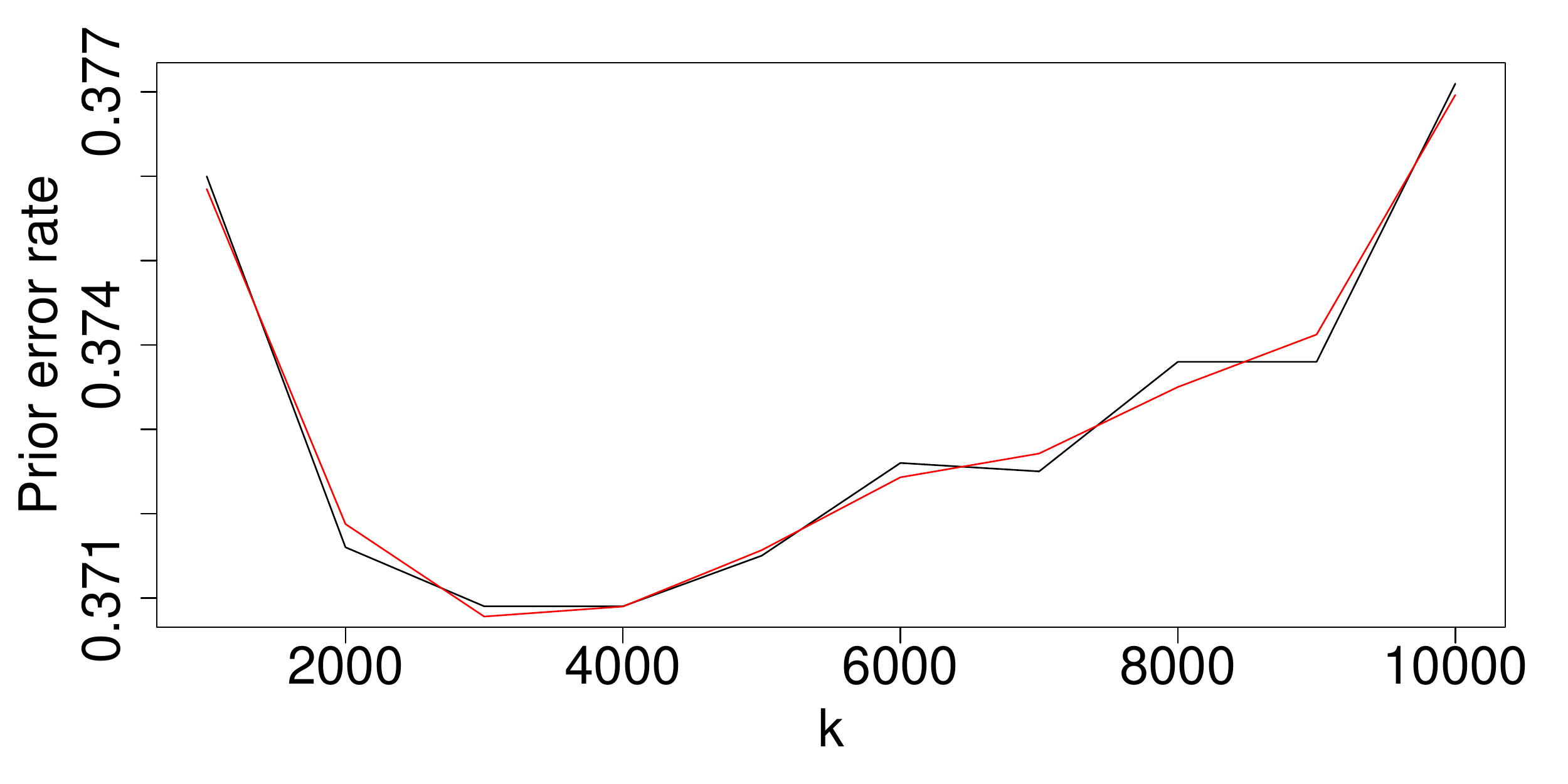}
  \caption{{\bf \sffamily \normalsize Calibration of \knn, the local logistic regression.} 
  \sffamily \normalsize Plot of the empirical prior error
  rate (\textit{in black}) of three classifiers, namely \knn\ (\textit{left}), the local logistic regression
  (\textit{right}) as a function of their tuning parameter when analyzing the Harlequin ladybird data with
  a reference table of $10,000$ simulations. To remove the noise of these estimated errors on a validation
  set composed of $10,000$ independent simulations, estimated errors are smoothed by a spline method that produces the \textit{red}
  curve. The optimal values of the parameters are $k=300$ and $k=3,000$ respectively. \label{fig:elbow}}
  \end{center}
\end{figurehere}

\section{A revealing toy example: MA(1) versus MA(2) models}

Given a time series $(x_t)$ of length $T=100$, we compare fits by moving average models of order either $1$ or
$2$, MA(1) and MA(2), namely
\[ 
x_t=\epsilon_t - \theta_1 \epsilon_{t-1} \ \text{and}\ x_t=\epsilon_t - \theta_1 \epsilon_{t-1} - \theta_2
\epsilon_{t-2}\,, \ \epsilon_t\sim\text{N}(0,\sigma^2)\,, 
\] 
respectively.  As previously suggested \cite{marin:pudlo:robert:ryder:2012}, a possible set of (insufficient)
summary statistics is made of the first seven autocorrelations, set that yields an ABC reference table
of size $N_\text{ref}=10^4$ with seven covariates. For both models, the
priors are uniform distributions on the stationarity domains \cite{robert:2001}:
\begin{itemize}
\item[--] for MA(1), the single parameter $\theta_1$ is drawn uniformly from the segment $(-1;1)$;
\item[--] for MA(2), the pair $(\theta_1,\theta_2)$ is drawn uniformly over the triangle defined by
  \[
  -2 < \theta_1 < 2,\quad \theta_1+\theta_2 >1,\quad \theta_1-\theta_2<1.
  \]
\end{itemize}

In this example, we can evaluate the discrepancy between the posterior probabilities based on the whole data
and those based on summaries. We first consider as summary statistics the first two
autocorrelations that are very informative for this problem.
The marginal likelihoods based on the whole data can be computed by numerical integrations of dimension $1$ and
$2$ respectively, while the ones based on the summary statistics are derived from a well-adapted
kernel smoothing method. Figure 1 of the main text shows how different the
posterior probabilities are when based on (i) the whole series of length $T=100$ and (ii) only the first
two autocorrelations, even though the latter remain informative about the problem. This graph brings numerical
support to the severe warnings of \cite{robert:cornuet:marin:pillai:2011}.

Moreover, Table~\ref{tab:MAMA} draws a comparison between various classifiers
when we consider as summary statistics the first seven autocorrelations. All methods based on
summaries are outperformed by the Bayes classifier that can be computed here via approximations of
the genuine $\pi(m|x)$: this ideal classifier achieves a prior error of around $12\%$. 
The RF classifier achieves the minimum with a prior error rate of around $16\%$.

If we now turn to the performances of the second random forest to evaluate the posterior probability
of the selected model (computed with Algorithm~3 displayed in the main text), Figure \ref{fig:MAMA.post}
shows how the evaluation of the posterior probability of the model selected by Algorithm~3 vary when
compared to the true posterior probability of the selected model. When the true posterior
probability of the selected model is high, Algorithm~3 has a tendency to underestimate the
probability.  Another important feature is that this approximation of the posterior probability do
not provide any warning regarding a decision swap between the true MAP model and the model selected
by Algorithm~2: see the black dots of Figure \ref{fig:MAMA.post}.

\newpage

\begin{tablehere}
\begin{center}
\begin{tabular}{cr}
        \hline
        \textbf{Classification method}                         & \textbf{Prior error rate (\%)} \\
        Linear discriminant analysis (LDA)                     & $26.57$ \\
        Logistic regression                                    & $27.40$ \\
        Na\"ive Bayes                                          & $24.40$ \\
        \knn\ with $k=100$ neighbors                           & $18.37$ \\
        \knn\ with $k=50$ neighbors                            & $17.35$ \\
        Random forests                                         & $16.15$ \\
        \hline
      \end{tabular}
\end{center}
  \caption{{\bf \sffamily \normalsize Estimated prior error rates in the MA(1) vs. MA(2) example.} 
      The prior error rates displayed here were computed as averaged misclassification
      errors on a set of $10^4$ simulations independent of the simulation set of $10^4$ values that
      trained the classifiers. Summary statistics are the first seven autocorrelations.
      A baseline error of $12.36\%$ is obtained when comparing the genuine posterior probabilities 
      on the whole data.\label{tab:MAMA}}
\end{tablehere}

\vspace{2cm} \begin{figurehere}
\centerline{\includegraphics[width=0.45\textwidth]{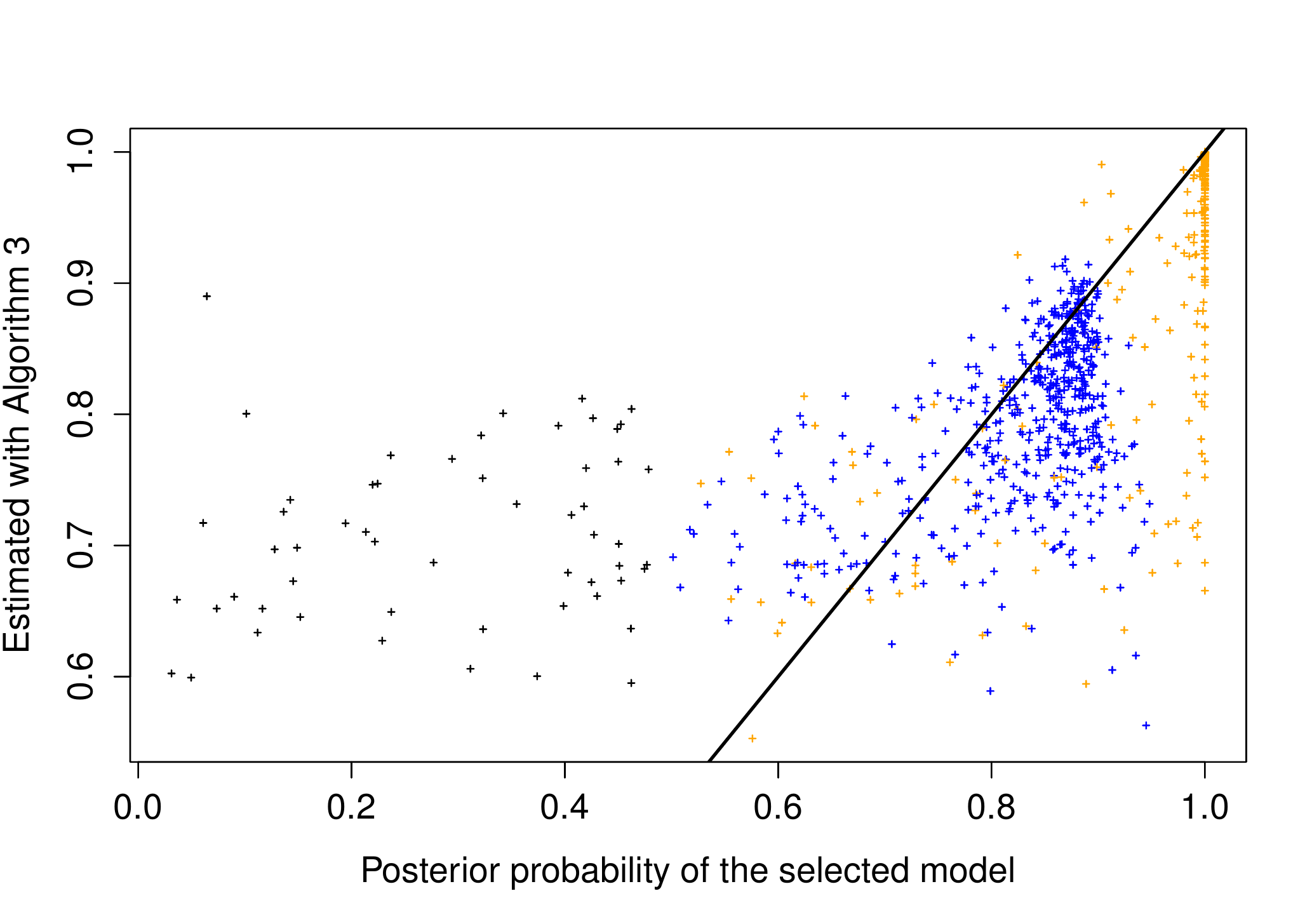}}
  \caption{{\bf \sffamily \normalsize Posterior probability of the selected model in the MA(1)
      vs. MA(2) example.} The graph compares the true posterior probabilities (\textit{x-axis}) of
    the decision taken by Algorithm 2 on 1,000 simulated time series with the approximation of
    these probabilities (\textit{y-axis}) provided by Algorithm 3. For each time series, the
    decision taken by Algorithm 2 can agree with the true model index that has MAP probability: dots
    are \textit{blue} when both are in favor of the MA(1) model, \textit{orange} when in favor of
    the MA(2) model. Black dots represents simulated time series which leads to a discrepancy
    between both decisions.\label{fig:MAMA.post}}
\end{figurehere}

\newpage

\section{Examples based on controlled simulated population \\ genetic datasets}

We now consider a basic population genetic setting ascertaining historical links between three populations
of a given species. In both examples below, we try to decide whether a third (and recent) population emerged
from a first population (Model 1), or from a second population that split from the first one some time ago
(Model 2), or whether this third population was a mixture between individuals from both populations (Model 3);
see Figure \ref{fig:scenarii}. 

\begin{figurehere}
\begin{center}
    \includegraphics[width=.3\textwidth]{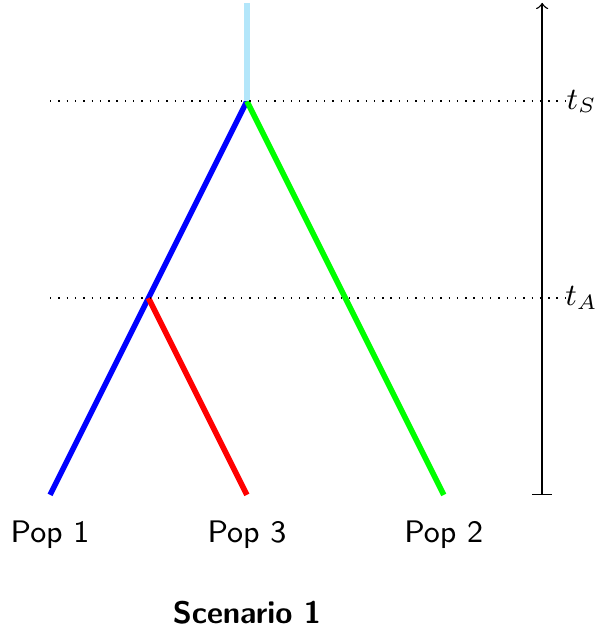}~\includegraphics[width=.3\textwidth]{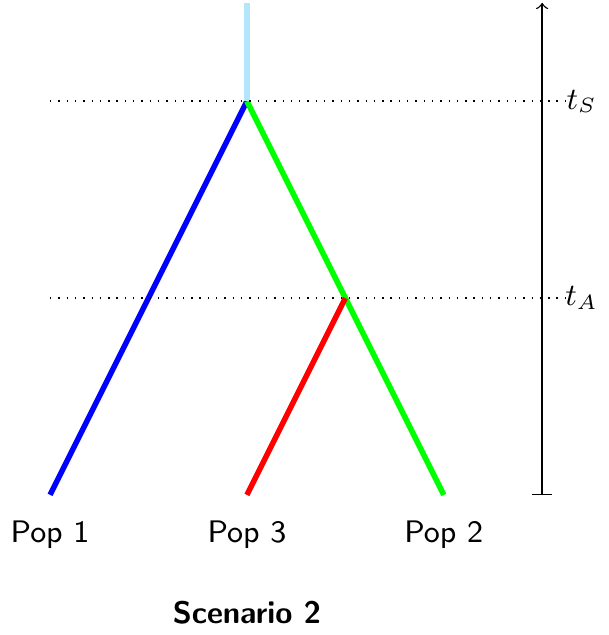}
    \includegraphics[width=.3\textwidth]{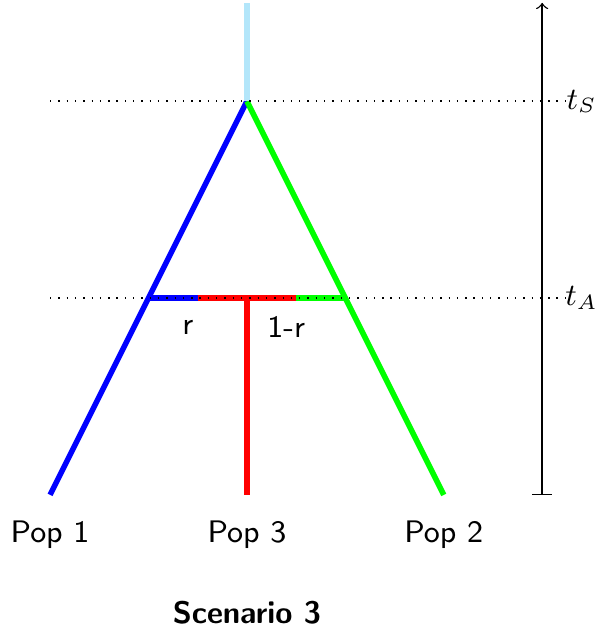}
  \end{center}
  \caption{{\bf \sffamily \normalsize Three competing models of historical relationships between populations of a
      given species.} \sffamily \normalsize Those population models or scenarios are used for both controlled
    examples based on SNP and microsatellite data: {\em (left)} Model 1 where Population 3 split from Population 1,
    {\em (center)} Model 2 where Population 3 split from Population 2, {\em (right)} Model 3 where Population 3 is
    an admixture between Populations 1 and 2. Branches in blue represent population of effective size
    $N_1$, in green $N_2$, in red $N_3$ and in cyan $N_4$.}
  \label{fig:scenarii}
\end{figurehere}

The only difference between both examples stands with  the kind of data they
consider: the $1,000$ genetic markers of the first example are autosomal, single nucleotide polymorphims (SNP)
loci and the $20$ markers of the second example are autosomal microsatellite loci. We assume that, in both
cases, the data were collected on samples of 25 diploid individuals from each population. Simulated and
observed genetic data are summarized with the help of a few statistics described in
Section G of the appendix. They are all computable with the DIYABC software
\cite{cornuet:etal:2014} that we also used to produce simulated datasets; see also Section F below.

For both examples, the seven demographics parameters of the Bayesian model are
\begin{itemize}
\item[--] $t_S$: time of the split between Populations~1 and 2,
\item[--] $t_A$: time of the appearance of Population~3, 
\item[--] $N_1$, $N_2$, $N_3$: effective population sizes of Populations~1, 2 and 3, respectively, below time $t_S$,
\item[--] $N_4$: effective population size of the common ancestral population above $t_S$ and
\item[--] $r$: the probability that a gene from Population 3 at time $t_A$ came from Population~1.
\end{itemize}
This last parameter $r$ is the rate of the admixture event at time $t_A$ and as such specific to Model 3. Note
that Model 3 is equivalent to Model 1 when $r=1$ and to Model 2 when $r=0$. But the
prior we set on $r$ avoids nested models. Indeed, the prior distribution is as follows:
\begin{itemize}
\item[--] the times $t_S$ and $t_A$ (on the scale of number of generations) are drawn from a uniform
  distribution over the segment $[10;\ 3\times 10^4]$ conditionally on $ t_A<t_S$;
\item[--] the four effective population sizes $N_i$, $i=1,\ldots,4$ are drawn independently from a uniform
  distribution on a range from $100$ to $30,000$ diploid individuals, denoted $U(100; 3\times 10^4)$;
\item[--] the admixture rate $r$ is drawn from a uniform distribution $U(0.05;\ 0.95)$.
\end{itemize}
In this example, the prior on model indices is uniform so that each of the three models has a prior probability of $1/3$.

\subsection{SNP data}\label{sub:SNP}

The data is made of $1,000$ autosomal SNPs for which we assume that the distances between these loci on the genome
are large enough to neglect linkage disequilibrium and hence consider them as having independent ancestral genealogies.
We use all summary statistics offered by the DIYABC software for SNP markers \cite{cornuet:etal:2014}, namely 48
summary statistics in this three population setting (provided in Section G of the appendix).  In total, we simulated
$70,000$ datasets, based on the above priors.
These datasets are then split into three groups:
\begin{itemize}
\item[--] $50,000$ datasets constitute the reference table and reserved for training classification steps,
  (we will also consider classifiers trained on subsamples of this set), 
\item[--] $10,000$ datasets constitute the validation set, used to calibrate the tuning parameters of the
  classifiers if needed, and
\item[--] $10,000$ datasets constitute the test set, used to evaluating the prior error rates.
\end{itemize}

The classification methods applied here are given in Table~\ref{tab:snp}.  For the na{\"i}ve Bayes classifier and
the LDA procedures, there is no parameter to calibrate. This is also the case of the RF methods
that, for such training sample sizes, do not require any calibration for $N_\text{boot}$.
The numbers $k$ of neighbors for the standard ABC techniques and for the local logistic regression
are tuned as described in Section A above. 

The prior error rates are estimated and minimized by using the validation set of $10^4$ simulations, independent from
the reference table.  The optimal value of $k$ for the standard ABC (\knn) and 48 summary statistics is small
because of the dimension of the problem ($k=9$, $k=15$, and $k=55$ when using $10,000$, $20,000$ and $50,000$
simulations in the reference table respectively). The optimal values of $k$ for the local logistic regression are
different, since this procedure fits a linear model on weighted neighbors. The calibration on a validation set made
of $10,000$ simulations produced the following optimal values: $k=2,000$, $k=3,000$, and $k=6,000$ when fitted on
$10,000$, $20,000$, and $50,000$ simulations, respectively. As reported in Section A above,
calibrating the parameter $k$ of the local logistic regression is very time consuming.  For the standard ABC (\knn) based on original summaries, we relied on a standard Euclidean distance
after normalizing each variable by its median absolute deviation, while \knn\ on the LDA axes requires no normalization procedure.

Table~\ref{tab:snp} provides estimated prior error rates for those classification techniques, based on a test
sample of $10,000$ values, independent of reference tables and calibration sets.  It shows that
the best error rate is associated with a RF trained on both the original DIYABC statistics and the LDA
axes. The gain against the standard ABC solution is clearly significant.  Other interesting features
exhibited in Table~\ref{tab:snp} are {(i)} good performances of the genuine LDA method, due to a good
separation between summaries coming from the three models, as exhibited in Figure \ref{fig:snp_lda}, albeit
involving some overlap between model clusters, {(ii)} that the local logistic regression on the two LDA axes of
\cite{estoup:etal:2012} achieves the second best solution.

Figure \ref{fig:snp_viss} describes further investigations into the RF solution.  This graph expresses the
contributions from the summary statistics to the decision taken by RF. The contribution of each summary is
evaluated as the average decrease in the Gini criterium over the nodes driven by the corresponding summary
statistic, see e.g. Chapter 15 of \cite{hastie:tibshirani:friedman:2009}. The appeal of including the first
two LDA axes is clear in Figure \ref{fig:snp_viss}, where they appear as LD1 and LD2: those statistics contribute
more significantly than any other statistic to the decision taken by the classifier. Note that the FMO
statistics, which also have a strong contribution to the RF decisions, are the equivalent of pairwise
$F_{ST}$-distances between populations when genetic markers are SNPs. The meaning of
the variable acronyms is provided in Section G below.

We simulated two typical datasets,
hereafter considered as pseudo-observed datasets or pod(s).  The first pod (green star in Figure
\ref{fig:snp_lda}) corresponds to a favorable situation for which Model 3 should easily be 
discriminated from both Models 1 and 2. The parameter values used to simulate this pod indeed correspond to a
recent balanced admixture between strongly differentiated source populations ($N_1=20,000$, $N_2=15,000$,
$N_3=10,000$, $N_4=25000$, $r=0.5$, $t_a=500$ and $t_s=20000$).  The second pod (red star in Figure
\ref{fig:snp_lda}) corresponds to a less favorable setting where it is more difficult to discriminate Model 3
from Model 1 and 2.  The parameter values used to simulate this second pod correspond to an ancient unbalanced
admixture between the source populations ($N_1=20,000$, $N_2=15,000$, $N_3=10,000$, $N_4=25,000$, $r=0.1$,
$t_a=10,000$, and $t_s=20,000$).

For both pods, ABC-RF (trained on both the 48 initial statistics and the two LDA axes) chooses Model 3. The RF
was trained on a reference table of size $70,000$ (that contains all simulations). The first pod is allocated
to Model 3 for all the 500 trees in the forest while the second pod is allocated to Model 3 for
261 trees (238 for Model 2 and only 1 for Model 1). As already explained, these numbers are very variable and does not
represent valid estimations of the posterior probabilities. The regression RF evaluation of the 
posterior probabilities of the selected model (Model 3), based on 500 trees and using the 0-1 error vector
(of size 70,000), are substantially different for both pods: very close to $1$ for the first one and about
$0.756$ for the second.
These posterior probabilities can be compared to the $17.9\%$ out-of-bag estimation of 
the prior error rate. Indeed, the first pod is attributed to Model 3 with an error close to zero
and the second one  with an error about $24.6\%$. That is totally in agreement with the position
of the pods represented by the green and red stars in Figure \ref{fig:snp_lda}.
The second pod belongs to a region of the data space where it seems difficult to
discriminate between Models 2 and 3.

\newpage

\subsection{Microsatellite data}\label{sub:microsat}

This illustration reproduces the same settings as in the SNP data example above but the genetic data (which
is of much smaller dimension) carries a different and lower amount of information. Indeed, we consider here
datasets composed of only 20 autosomal microsatellite loci. The microsatellite loci are assumed to follow a
generalized stepwise mutation model with three parameters \cite{estoup:jarne:cornuet:2002,lombaert:etal:2011}:
the mean mutation rate ($\bar \mu$), the mean parameter of the geometric distribution
($\bar P$) of changes in number of repeats during mutation events, and the mean mutation rate for single
nucleotide instability ($\overline{\mu_{SNI}}$). The prior distributions for $\bar \mu$, $\bar P$ and
$\overline{\mu_{SNI}}$ are the same as those given in Table~\ref{tab:asian} (see the prior distributions used
for the real \textit{Harmonia axyridis} microsatellite dataset). Each locus has a possible range of 40 contiguous
allelic states and is characterized by locus specific $\mu$'s drawn from a Gamma distribution with mean 
$\bar \mu$ and shape $2$, locus specific $P$'s drawn from a Gamma distribution with mean 
$\bar P$ and shape $2$ and, finally, locus specific $\mu_\text{SNI}$'s drawn from a Gamma
distribution with mean $\overline{\mu_{SNI}}$ and shape $2$. For microsatellite markers,
DIYABC \citep{cornuet:etal:2014} produces 39 summary statistics described in Section G below.

Table~\ref{tab:microsat} is the equivalent of Table~\ref{tab:snp} for this kind of genetic data structure. Due
to the lower and different information content of the data, the prior error rates are much higher in all
cases, but the conclusion about the gain brought by RF using all summaries plus the LDA statistics remains.

As in the SNP case, we simulated two typical pods: one highly favorable (the green star in Figure \ref{fig:microsat_lda})
and a second one quite challenging (the red star in Figure \ref{fig:microsat_lda}).  They were generated using the same
values of parameters as for the SNP pods.

For both pods, we considered an ABC-RF treatment with a reference table of size $70,000$.
The first pod is allocated to Model 3: 491 trees associate this dataset to Model 3, 
3 to Model 1 and 6 to Model 2. The second pod is also allocated to Model 3:
234 trees associate this dataset to Model 3, 61 to Model 1 and 205 to Model 2.
We run a regression RF model to estimate the corresponding posterior probabilities.
Using once again 500 trees and the $70,000$ 0-1 error vector we obtained
about $0.996$ for the first pod and about $0.598$ for the more challenging one.
We hence got  for the challenging pod a
classification error (1 minus the posterior probability of the selected model)
that is larger than the prior error rate.

Interestingly Figure \ref{fig:micro_viss} shows that the AML\_3\_1\&2 summary statistic (see Section G below) contributes
more to the RF decision than the second LDA axis. We recall that AML is an admixture rate estimation computed
by maximum likelihood on a simplified model considering that the admixture occurred at time $t=0$. The
importance of the LDA axes in the random forest remains nevertheless very high in this setting.

\newpage 

\begin{figurehere}
\centerline{\includegraphics[width=0.35\textwidth]{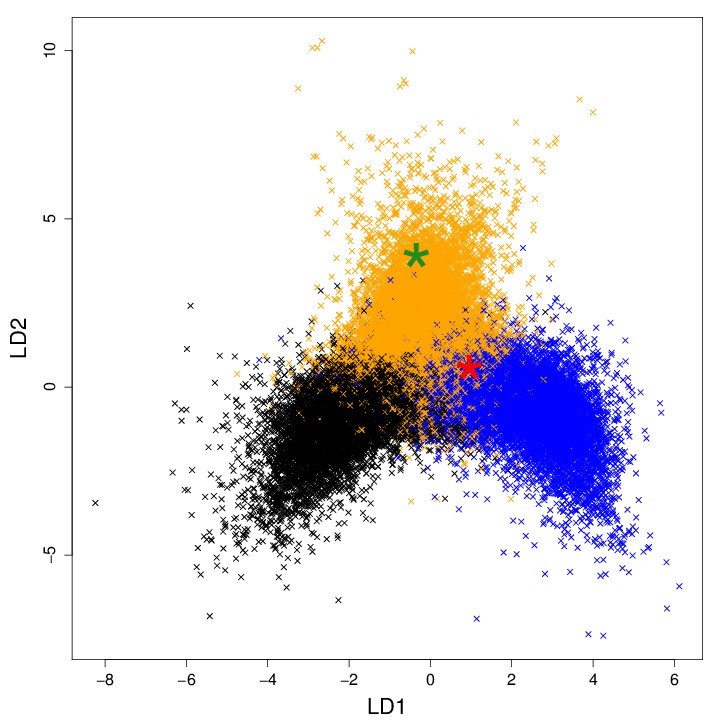}}
    \caption{{\bf \sffamily \normalsize Projections on the LDA axes of the simulations from the reference
        table for the controlled SNP example.}  \sffamily \normalsize Colors correspond to model indices: black for Model 1, blue for
      Model 2 and orange for Model 3. The locations of both simulated pseudo-observed datasets that are analyzed as
      if they were truly observed data, are indicated by {\em green and red stars}.  }
     \label{fig:snp_lda}
\end{figurehere}

\vspace{2cm} \begin{figurehere}
  \begin{center}
  \includegraphics[width=0.45\textwidth]{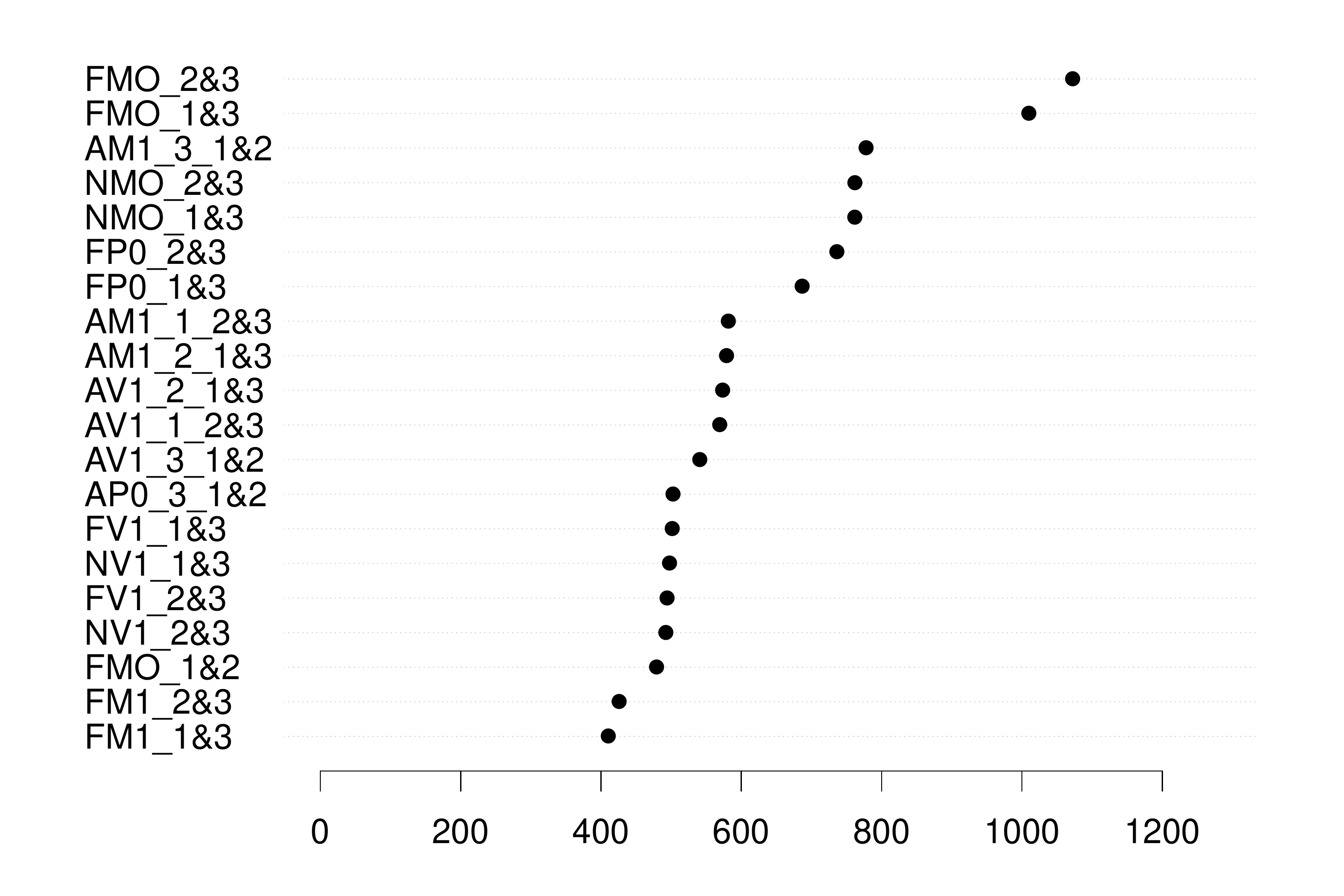}
  \includegraphics[width=0.45\textwidth]{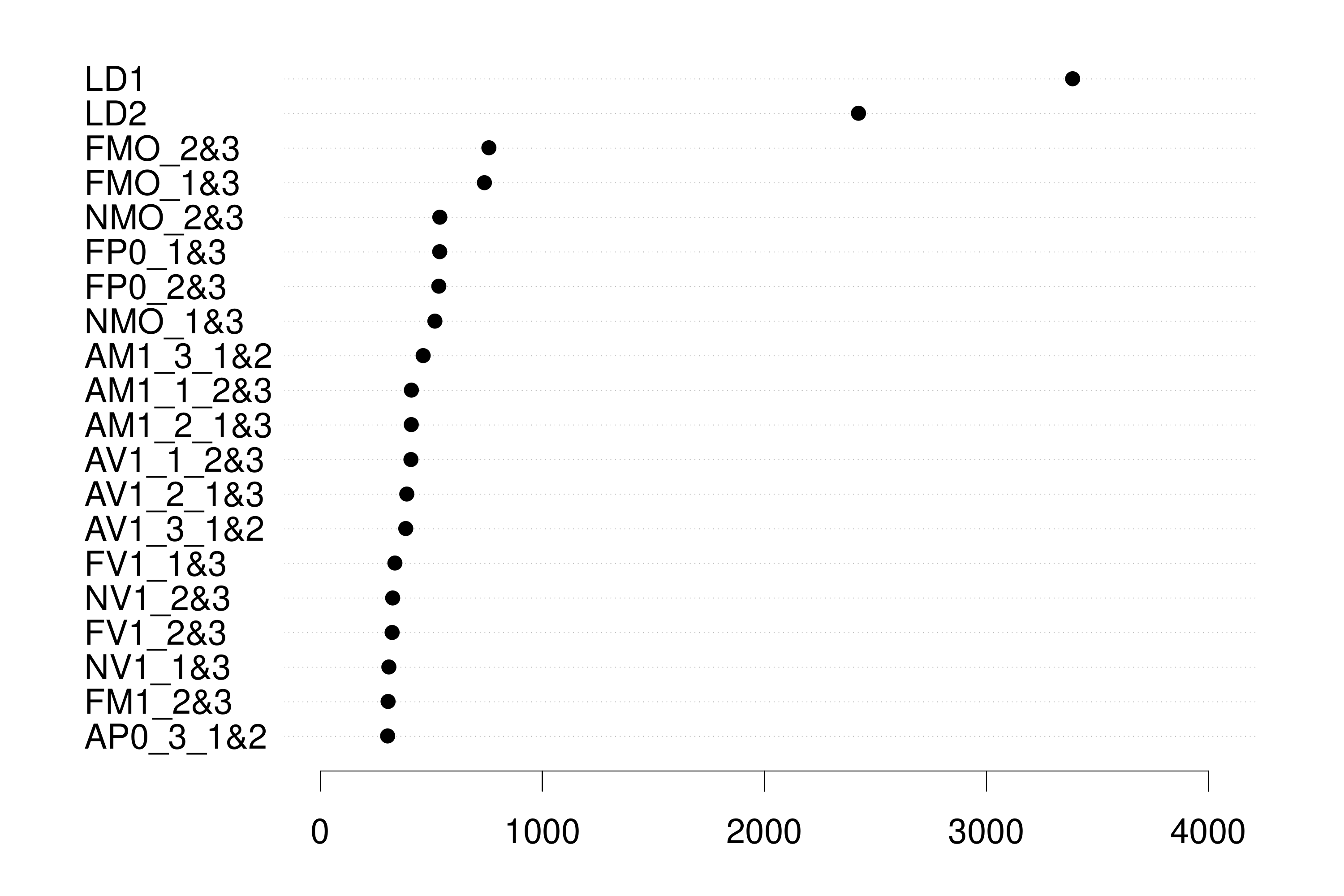}
  \end{center}
  \caption{{\bf \sffamily \normalsize Contributions of the twenty most important statistics to the RF for the
  controlled SNP example.} \sffamily \normalsize The contribution of a statistic is evaluated as the
    mean decrease in the Gini criterium when using 48 summary statistics {\em (top)} and
     when adding the two LDA axes to the previous set of statistics {\em (bottom)}. \label{fig:snp_viss}}
\end{figurehere}

\newpage

\begin{figurehere}
  \centering
  \includegraphics[width=0.35\textwidth]{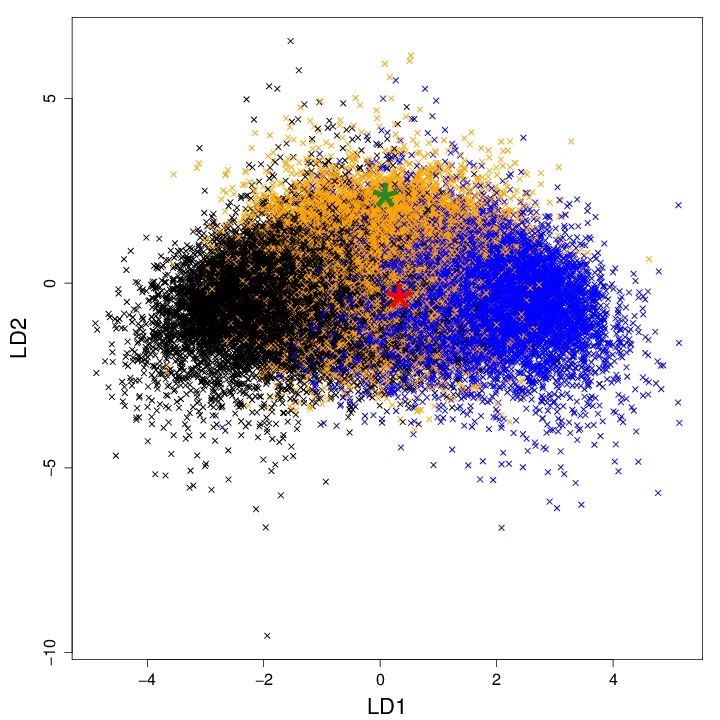}
  \caption{{\bf \sffamily \normalsize Projections on the LDA axes of the simulations from the reference
      table for the controlled microsatellite example} \sffamily \normalsize Colors correspond to model indices: black for Model 1, blue for Model
    2 and orange for Model 3.  The locations of both simulated pseudo-observed datasets are indicated by {\em green and
      red stars}.  }
  \label{fig:microsat_lda}
\end{figurehere}

\vspace{2cm} \begin{figurehere}
  \centering
  \includegraphics[width=0.45\textwidth]{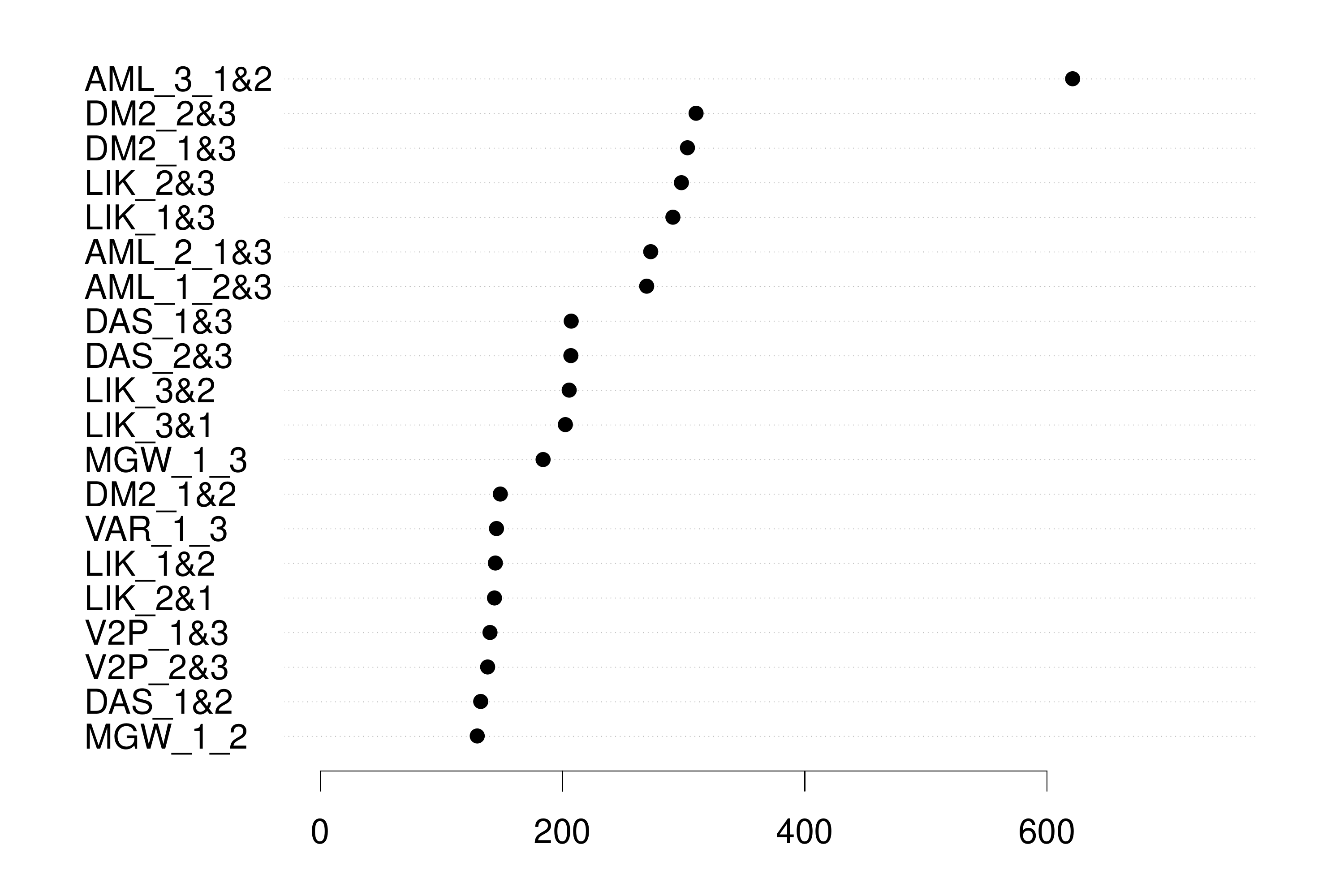}
  \includegraphics[width=0.45\textwidth]{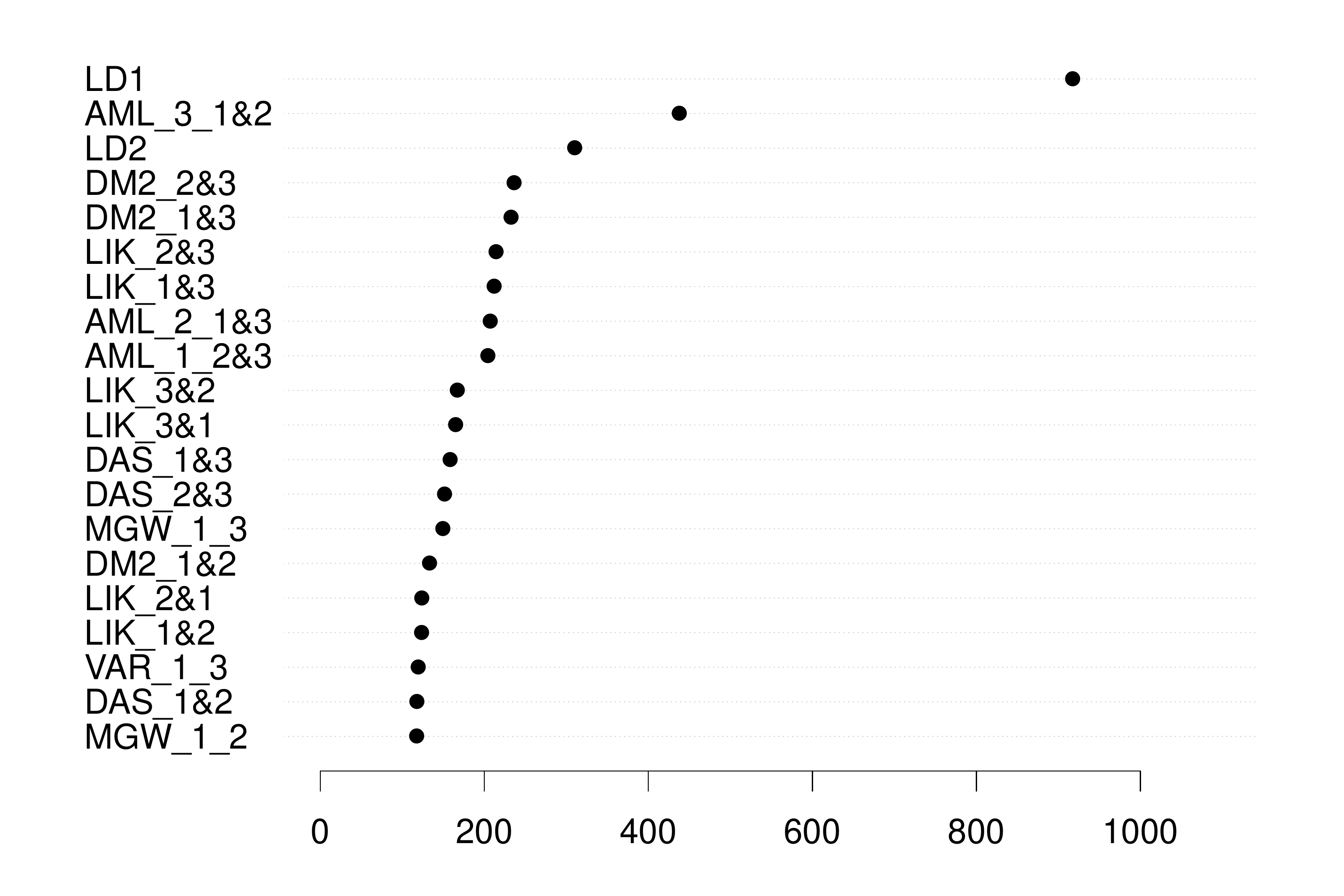}
    \caption{{\bf \sffamily \normalsize Contributions of the twenty most important statistics to the RF for the
        controlled microsatellite example.} \sffamily \normalsize The contribution of a statistic is evaluated with the
    mean decrease in the Gini criterium when using 39 summary statistics {\em (top)} and
     when adding the two LDA axes to the previous set of statistics {\em (bottom)}.\label{fig:micro_viss}}
\end{figurehere}

\newpage 

{\small \begin{tablehere}
    \begin{center}
    \begin{tabular}{crrr}
      \hline 
      \textbf{Classification method } & 
      \multicolumn{3}{c}{\textbf{Prior error rates} (\%) } \\ 
      \textbf{trained on} & $N_\text{ref}=10,000$ & $N_\text{ref}=20,000$ & $N_\text{ref}=50,000$ \\
      \hline
      Na\"ive Bayes (with Gaussian marginals)           & $34.86$ & $34.72$ & $34.41$ \\
      Linear discriminant analysis (LDA)                & $23.35$ & $23.28$ & $23.12$ \\
      Standard ABC (\knn) using DIYABC summaries        & $27.82$ & $25.61$ & $23.58$ \\
      Standard ABC (\knn) using only LDA axes           & $22.86$ & $22.56$ & $22.60$ \\ 
      Local logistic regression on LDA axes             & $22.07$ & $21.84$ & $21.93$ \\
      RF using DIYABC initial summaries                 & $22.13$ & $20.71$ & $19.55$ \\
      RF using both DIYABC summaries and LDA axes 		& $20.07$ & $18.95$ & $18.11$ \\
      \hline
    \end{tabular}
  \end{center}
  \caption{{\bf \sffamily \normalsize Estimated prior error rates for the controlled SNP example.}\label{tab:snp}}
\end{tablehere}

\vspace{2cm} \begin{tablehere}
  \begin{center}
    \begin{tabular}{crrr}
      \hline 
      \textbf{Classification method } & 
      \multicolumn{3}{c}{\textbf{Prior error rates} (\%) } \\ 
      \textbf{trained on} & $N_\text{ref}=10,000$ & $N_\text{ref}=20,000$ & $N_\text{ref}=50,000$ \\
      \hline
      Na\"ive Bayes (with Gaussian marginals)                & $40.41$ & $40.74$ & $40.43$ \\
      Linear discriminant analysis (LDA)                     & $36.54$ & $36.49$ & $36.47$ \\
      Standard ABC (\knn) using DIYABC summaries             & $39.23$ & $38.37$ & $36.99$ \\
      Standard ABC (\knn) using only LDA axes                & $37.17$ & $36.93$ & $36.05$ \\ 
      Local logistic regression on LDA axes                  & $36.11$ & $35.93$ & $35.89$ \\
      RF using DIYABC initial summaries                      & $37.07$ & $36.16$ & $35.32$ \\
      RF using both DIYABC summaries and LDA axes            & $35.53$ & $35.32$ & $34.38$ \\
      \hline
    \end{tabular}
  \end{center}
  \caption{{\bf  \sffamily \normalsize Estimated prior error rates for the controlled microsatellite example.}\label{tab:microsat}}
\sffamily \normalsize 
\end{tablehere}}

\newpage

\section{Supplementary informations about \\ the Harlequin ladybird example}

Reconstructing the history of the invasive populations of a given species is crucial for both management and
academic issues \cite{estoup:guillemaud:2010}. The real dataset analyzed here and in the main text relates to
the recent invasion history of a coccinellidae, \textit{Harmonia axyridis}. Native from Asia, this insect
species was recurrently introduced since 1916 in North America as a biocontrol agent against aphids. It
eventually survived in the wild and invaded four continents. The present dataset more specifically aims at
making inference about the introduction pathway of the invasive \textit{H. axyridis} for the first recorded
outbreak of this species in eastern North America, which corresponds to a key introduction event in the
worldwide invasion history of the species. We refer the reader to \cite{estoup:guillemaud:2010} and
\cite{lombaert:etal:2011} for more details on the biological issues associated to the situation considered and
for a previous statistical analysis based on standard ABC techniques.

Using standard ABC treatments, \cite{lombaert:etal:2011} formalized and compared ten different scenarios (i.e.,
models) to identify the source population(s) of the eastern North America invasion (see
Figure \ref{fig:asian.winscen} for a graphical representation of a typical invasion scenario and
Table~\ref{tab:prior} for parameter prior distributions). We now compare our results based on the ABC-RF algorithm
with this original paper, as well as with other classification methods. The \textit{H. axyridis} dataset is made of
samples from five populations comprising 35, 18, 34, 26 and 25 diploid individuals, genotyped at 18 autosomal
microsatellite loci considered as selectively neutral and statistically independent markers. The problem we face is
considerably more complex than the above controlled and simulated population genetic illustrations in that the
numbers and complexity levels of both competing models and sampled populations are noticeably higher. Since the
summary statistics proposed by DIYABC \cite{cornuet:etal:2014} describe features that operate per population, per
pair, or per triplet of populations, averaged over the 18 loci, we can include up to 130 of those statistics plus
the nine LDA axes as summary statistics in our ABC-RF analysis. The gap in the dimension of the summary statistics
is hence major when compared with the 48 and 39 sizes in the previous sections. The ABC-RF computations to
discriminate among the ten scenarios and evaluate error rates were processed on 10,000, 20,000, and 50,000 datasets
simulated with DIYABC; see Section~7 below and \cite{cornuet:etal:2014}. It is worth noting here that the standard
ABC treatments processed in \cite{lombaert:etal:2011} on the same dataset relied on a set of 86 summary statistics
which were selected thanks to the author expertise in population genetic model choice, as well as on a
substantially larger number of simulated datasets (i.e., $1,000,000$ per scenario).
Compared to the standard ABC treatments processed in \cite{lombaert:etal:2011}, ABC-RF computation allowed substantial 
computation speed: a factor 200 to 1000 for the simulation of the reference table.

\begin{figurehere}
    \centerline{\includegraphics[width=0.49\textwidth]{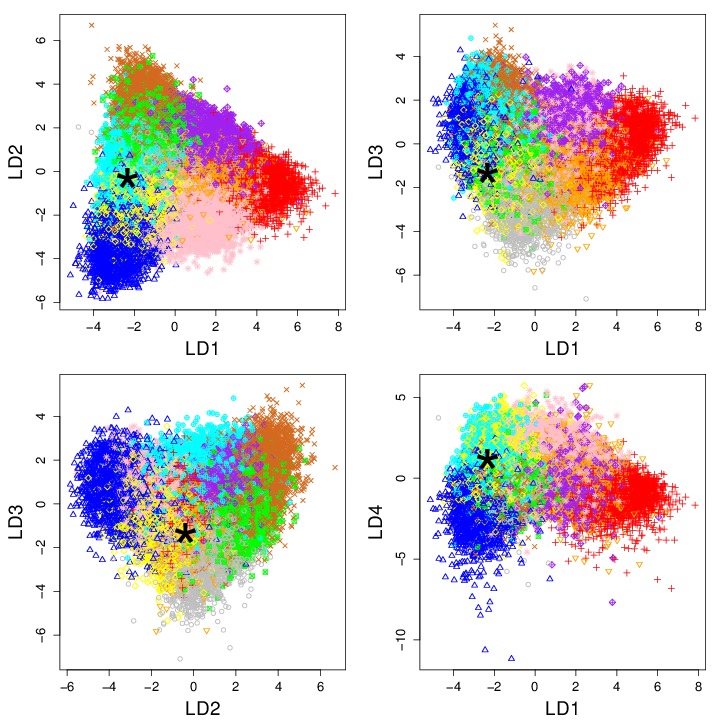}}\par

    \caption{{\bf \sffamily \normalsize Projections on the first four LDA axes of simulations from the reference
        table of the Harlequin ladybird analysis.}  \sffamily \normalsize Colors correspond to model indices. The
      location of the real observed dataset for the Harlequin ladybird is indicated by a \textit{black star}.
    \label{fig:cox_lda}}
\end{figurehere}

\vspace{2cm} \begin{figurehere}
  \centering
  \includegraphics[width=0.45\textwidth]{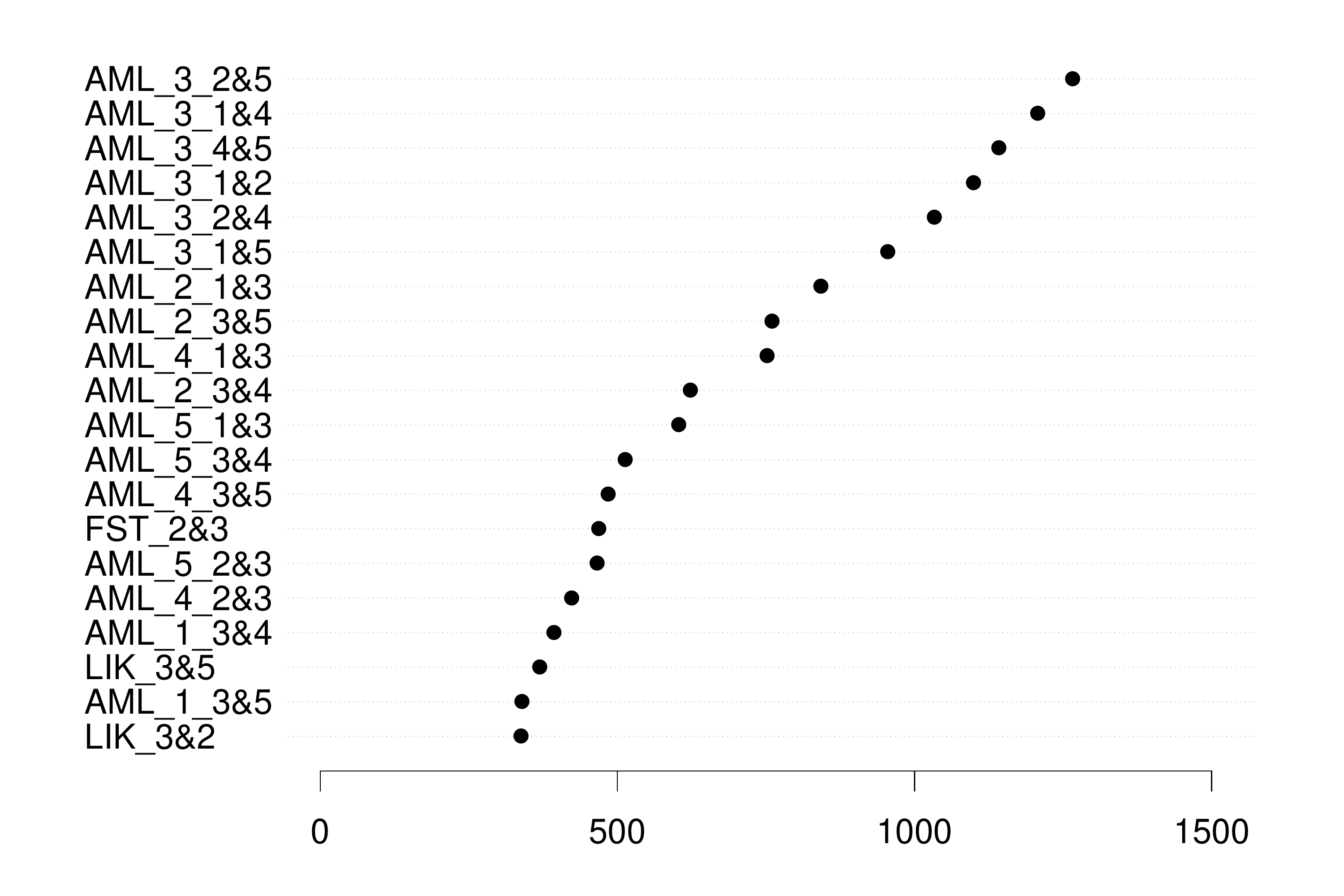} ~
  \includegraphics[width=0.45\textwidth]{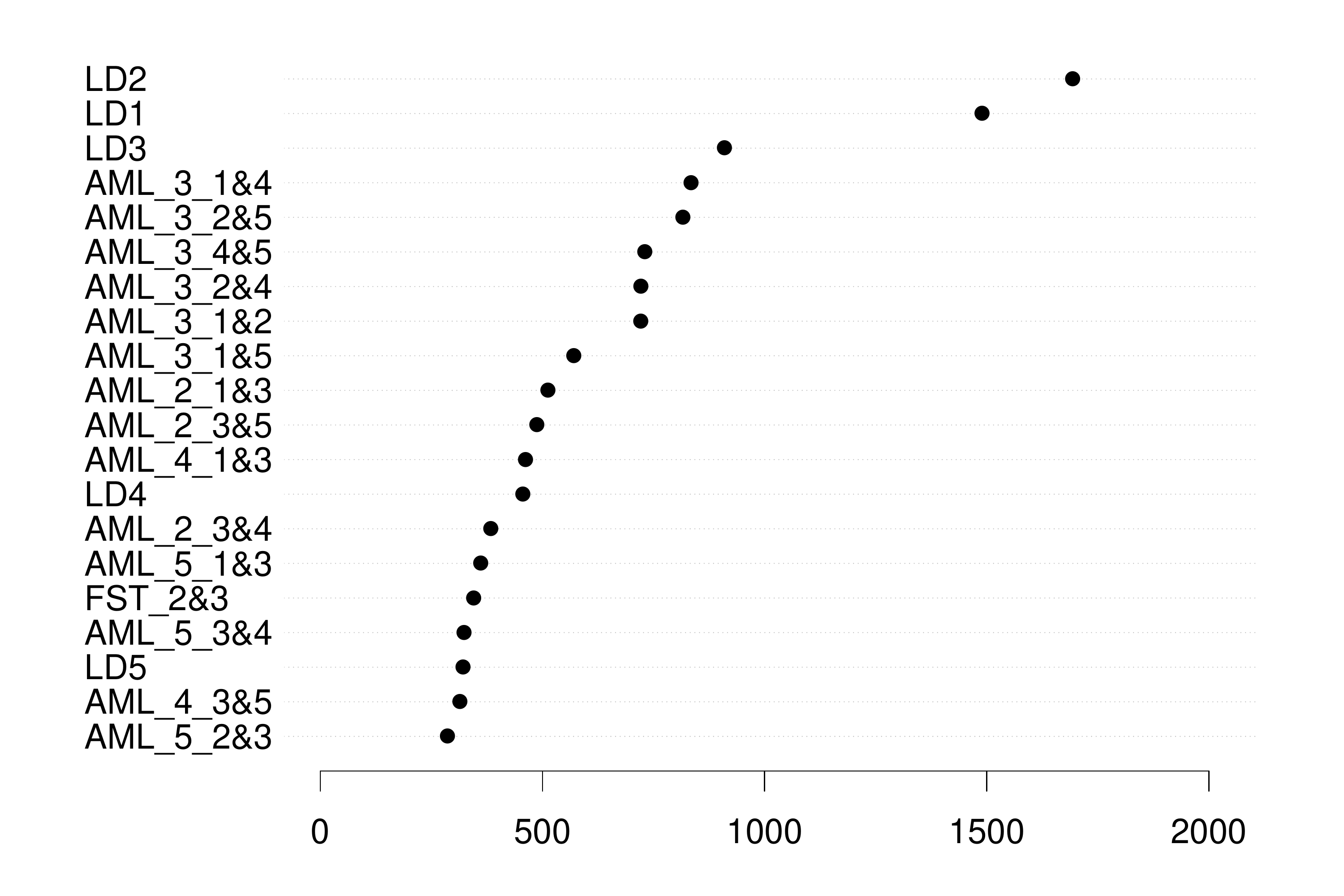}
  \caption{{\bf \sffamily \normalsize Contributions of the most important statistics to the RF for
      the Harlequin ladybird example.} \sffamily \normalsize The contribution of a statistic is evaluated with the
    mean decrease in node impurity in the trees of the RF when using 130 summary statistics {\em (top)} and
     when adding the nine LDA axes to the previous set of statistics {\em (bottom)}. The meaning of
    the variable acronyms is provided in Appendix S2.\label{fig:micro_boulon}}
  \label{fig:cox_viss}
\end{figurehere}

\begin{figurehere}

  \centerline{\includegraphics[width=.8\textwidth]{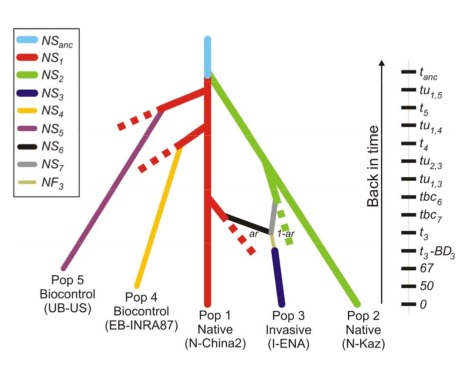}}\par

  \caption{{\bf \sffamily \normalsize Representation of the invasion scenario of the Harlequin
      ladybird that was selected among ten scenarios when using either the proposed ABC-RF method or
      the standard ABC analysis processed in \cite{lombaert:etal:2011}.}
    \sffamily \normalsize 
     Population names between brackets correspond to sample names in
    \cite{lombaert:etal:2011}. Populations (i.e., Pop) 1 and 2 are native populations; Pop 3 is the target
    invasive population for which one wants to make inference about the source population(s) (here Pop 3 is an
    admixture between the source populations Pop 1 and Pop 2); Pop 4 and 5 are biocontrol populations; Pop
    ubc6 and ubc7 are unsampled biocontrol populations. Time 0 is the sampling date of Pop 1, 2 and 3. Time 50
    and time 67 are the sampling dates of Pop 4 and 5 respectively. Dashed lines correspond to unsampled
    native populations; see \cite{lombaert:etal:2011} for a comprehensive explanation of this design. Populations
    were assumed to be isolated from each other, with no exchange of migrants. All parameters 
    are described with their associated prior in Table~\ref{tab:asian}. This scenario was opposed to nine competing scenarios
    corresponding to alternative single or multiple source(s) for the introduced target population Pop 3; see
    \cite{lombaert:etal:2011} for details.
    \label{fig:asian.winscen}}
\end{figurehere}

\begin{tablehere}
\begin{center}
\begin{tabular}{cc} 
\hline
\textbf{parameter} & \textbf{prior}  \\
\hline
$\text{NS}_i$, $\text{NS}_j$     & $\mathcal{U}(100,20000)$ \\
$\log(\text{NS}_k)$              & $\mathcal{U}(10,10^4)$ \\
$\log(\text{NF}_i)$              & $\mathcal{U}(2,10^4)$ \\
$\text{BD}_i$                    & $\mathcal{U}(0,5)$ \\
ar                               & $\mathcal{U}(0.1,0.9)$ \\
$t_i$, $t_k$                     & $\mathcal{U}(x_i,x_i+5)$ \\
$\log(tbc_i)$                    & $\mathcal{U}(t_i,93)$ \\
$\log(tu_{ji})$, $\log(tu_{jk})$ & $\mathcal{U}(tbc_i,3000)$ \\
$t_{\text{anc}}$                 & $\mathcal{U}(100,3000)$ \\
$\bar \mu$                       & $\mathcal{U}(10^{-5},10^{-3})$ \\
$\bar P$                         & $\mathcal{U}(0.1,0.3)$ \\
$\overline{\mu_\text{SNI}}$      & $\mathcal{U}(10^{-8},10^{-4})$ \\
\hline
\end{tabular}
\end{center}
\sffamily \normalsize
\caption{ \label{tab:prior} \sffamily \normalsize {\bf Prior distributions of demographic, historical, and mutation parameters used to
simulate microsatellite datasets and corresponding summary statistics under the
ten invasion scenarios of Harlequin ladybird.}
Prior distributions are the same as those used in \cite{lombaert:etal:2011} and \cite{estoup:etal:2012}.
Populations $i$ are invasive populations, clusters $j$ are native population clusters
(either Western Asia or Eastern Asia cluster) and populations $k$ correspond to biocontrol strains
(i.e., laboratory reared populations). Times were translated into numbers of generations running back in time
and assuming 2.5 generations per year, $NS$ stands for stable effective population size (number of diploid individuals);
$NF$ for effective number of founders during an introduction step lasting BD generation(s); $ar$ for admixture rate
(only for scenarios with admixture); $t_i$ for introduction date of invasive populations $i$ with limits $x_i$ fixed
from dates of first observation, assuming 2.5 generations per year; $t_k$ for sampling date in the native area of
biocontrol population $k$ with limits $x_i$ fixed from historical information, assuming 2.5 generations per year;
$tbc_i$ for creation date of unsampled biocontrol strain for Eastern and Western North America populations bounded
by the dates of the first observation of the invasive population (corresponding to a direct introduction into
the wild) and the number of generations from 1970, the starting date of a period of intense \textit{H. axyridis}
biocontrol activity in the USA; $tu_j$ for native cluster $j$ date of merging of the source unsampled native
population with the sampled native population (this parameter is included in the model in which the scenario
contains one or both native populations as possible source(s) of population $i$ or population $k$); $t_\text{anc}$
for the date of
the merging of the two native populations into an ancestral unsampled population (with condition $tu_j \le
t_\text{anc}$). See \cite{lombaert:etal:2011} and \cite{estoup:etal:2012} for additional details.}
\end{tablehere}

\newpage

\section{Supplementary informations about \\ the Human population example}

We here illustrate the potential of our ABC-RF algorithm for the statistical processing of massive single
nucleotide polymorphism (SNP) datasets, whose production is on the increase within the field of population
genetics.  To this aim, we analyzed a SNP dataset obtained from individuals originating from four Human
populations (30 unrelated individuals per population) using the freely accessible public 1000 Genome databases
({i.e.}, the vcf format files including variant calls available at
\verb+http://www.1000genomes.org/data+).  The goal of the 1000 Genomes Project is to find most genetic
variants that have frequencies of at least 1\% in the studied populations by sequencing many individuals
lightly (i.e., at a $4\times$ coverage).  A major interest of using SNP data from the 1000 Genomes Project
\cite{genome:project:2012} is that such data does not suffer from any ascertainment bias ({i.e.}, the
deviations from expected theoretical results due to the SNP discovery process in which a small number of
individuals from selected populations are used as discovery panel), which is a prerequisite when using the
DIYABC simulator of SNP data \cite{cornuet:etal:2014}.  The four Human populations included the Yoruba
population (Nigeria) as representative of Africa (encoded YRI in the 1000 genome database), the Han Chinese
population (China) as representative of the East Asia (encoded CHB), the British population (England and
Scotland) as representative of Europe (encoded GBR), and the population composed of Americans of African
ancestry in SW USA (encoded ASW).  The SNP loci were selected from the 22 autosomal chromosomes using the
following criteria: (i) all $30\times4$ analyzed individuals have a genotype characterized by a quality score
$(GQ)>10$ (on a PHRED scale), (ii) polymorphism is present in at least one of the $30\times4$ individuals in
order to fit the SNP simulation algorithm of DIYABC \cite{cornuet:etal:2014}, (iii) the minimum distance
between two consecutive SNPs is 1 kb in order to minimize linkage disequilibrium between SNPs, and (iv) SNP
loci showing significant deviation from Hardy-Weinberg equilibrium at a 1\% threshold
\cite{wigginton:cutler:abecasis:2005} in at least one of the four populations have been removed (35 SNP loci
involved).  After applying the above criteria, we obtained a dataset including 51,250 SNP loci scattered over
the 22 autosomes (with a median distance between two consecutive SNPs equal to 7 kb) among which 50,000 were
randomly chosen for applying the proposed ABC-RF methods.

In this application, we compared six scenarios ({i.e.}, models) of evolution of the four Human
populations genotyped at the above mentioned 50,000 SNPs.  The six scenarios differ from each other by one
ancient and one recent historical event: {\em (i)} A single out-of-Africa colonization event giving an ancestral
out-of-Africa population which secondarily splits into one European and one East Asia population lineage,
versus two independent out-of-Africa colonization events, one giving the European lineage and the other one
giving the East Asia lineage. The possibility of a second ancient ({i.e.}, $>100,000$ years) 
out-of-Africa colonization event through the Arabian peninsula toward Southern Asia has been suggested by
archaeological studies, e.g. \cite{rose:etal:2011};  {\em (ii)} The possibility (or not) of a recent genetic
admixture of Americans of African ancestry in SW USA between their African ancestors and individuals of
European or East Asia origins.

The six different scenarios as well as the prior distributions of the time event and effective population size
parameters used to simulate SNP datasets using DIYABC are detailed in Figure \ref{fig:outofAf}.  We stress here that
our intention is not to bring new insights into Human population history, which has been and is still studied in
greater details in a number of studies using genetic data, but to illustrate the potential of the proposed ABC-RF
methods for the statistical processing of large size SNP datasets in the context of complex evolutionary histories.
RF computations to discriminate among the six scenarios of Figure \ref{fig:outofAf} and evaluate error rates were
processed on 10,000, 20,000, and 50,000 simulated datasets. We used all summary statistics offered by the DIYABC
software for SNP markers \cite{cornuet:etal:2014} (see Section G below), namely 130 summary statistics
in this setting plus the five LDA axes as additional summary statistics.

\vspace{2cm} \begin{figurehere}
  \centering
  \includegraphics[width=0.45\textwidth]{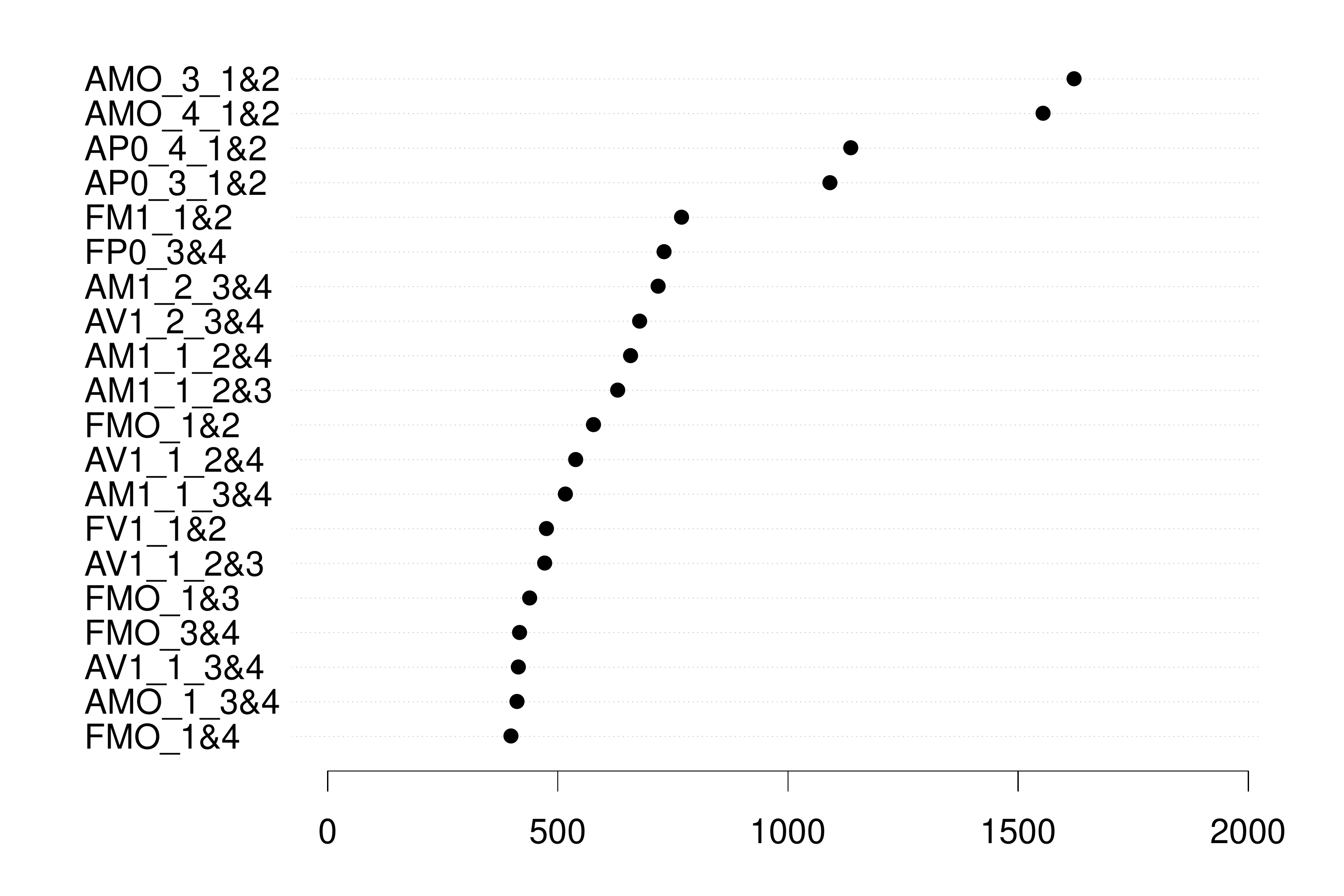}
  \includegraphics[width=0.45\textwidth]{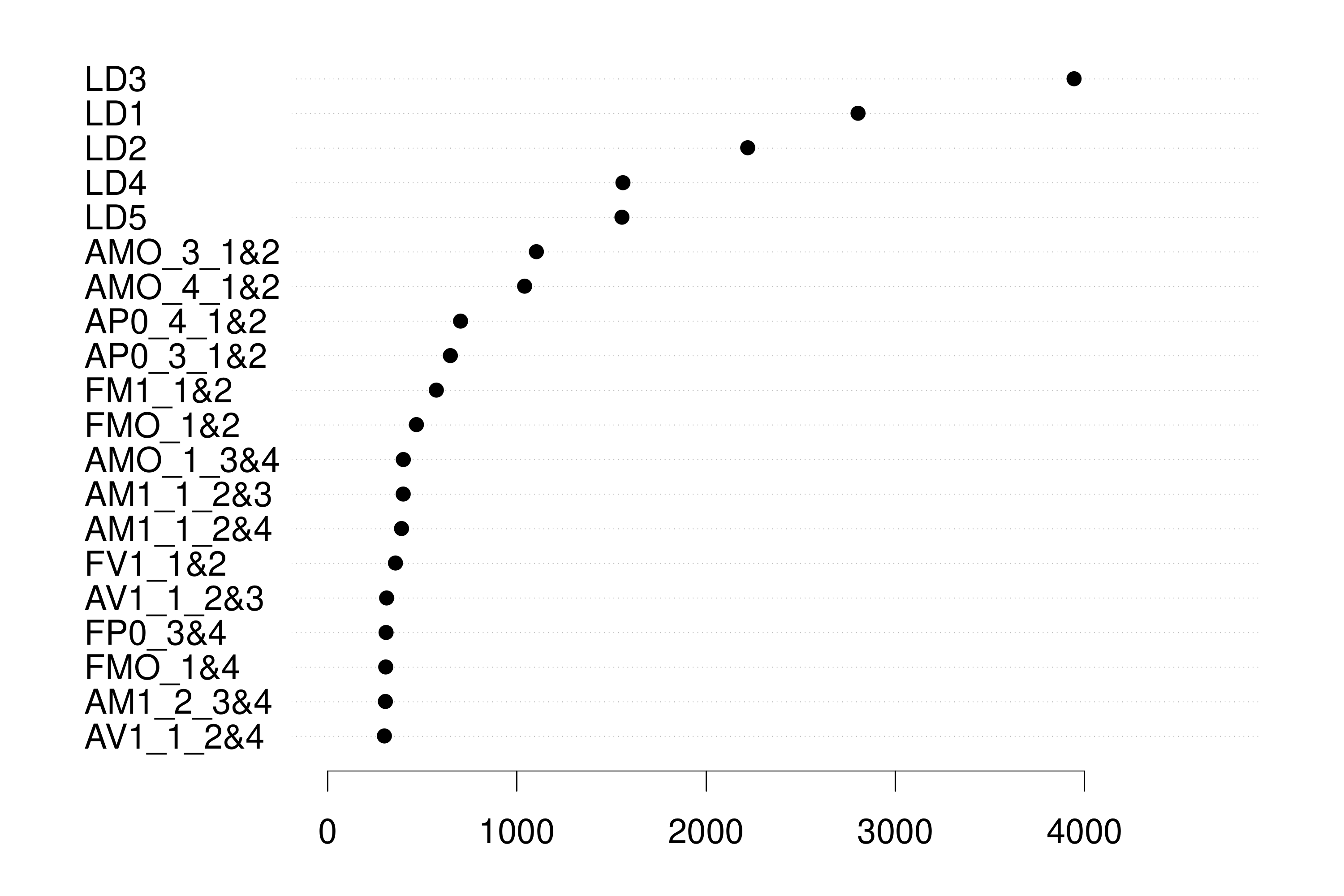}
    \caption{{\bf \sffamily \normalsize Contributions of the most important statistics to the RFs (to analyze
      the Human data).} \sffamily \normalsize The contribution of a statistic is evaluated with the
    mean decrease in node impurity in the trees of the RF when using the 112 summary statistics {\em (top)} and
     when adding the five LDA axes to the previous set of statistics {\em (bottom)} . The meaning of
    the variable acronyms is provided in Appendix S1.\label{fig:human_viss}}
\end{figurehere}

\begin{figurehere}
\centerline{\includegraphics[width=.6\textwidth]{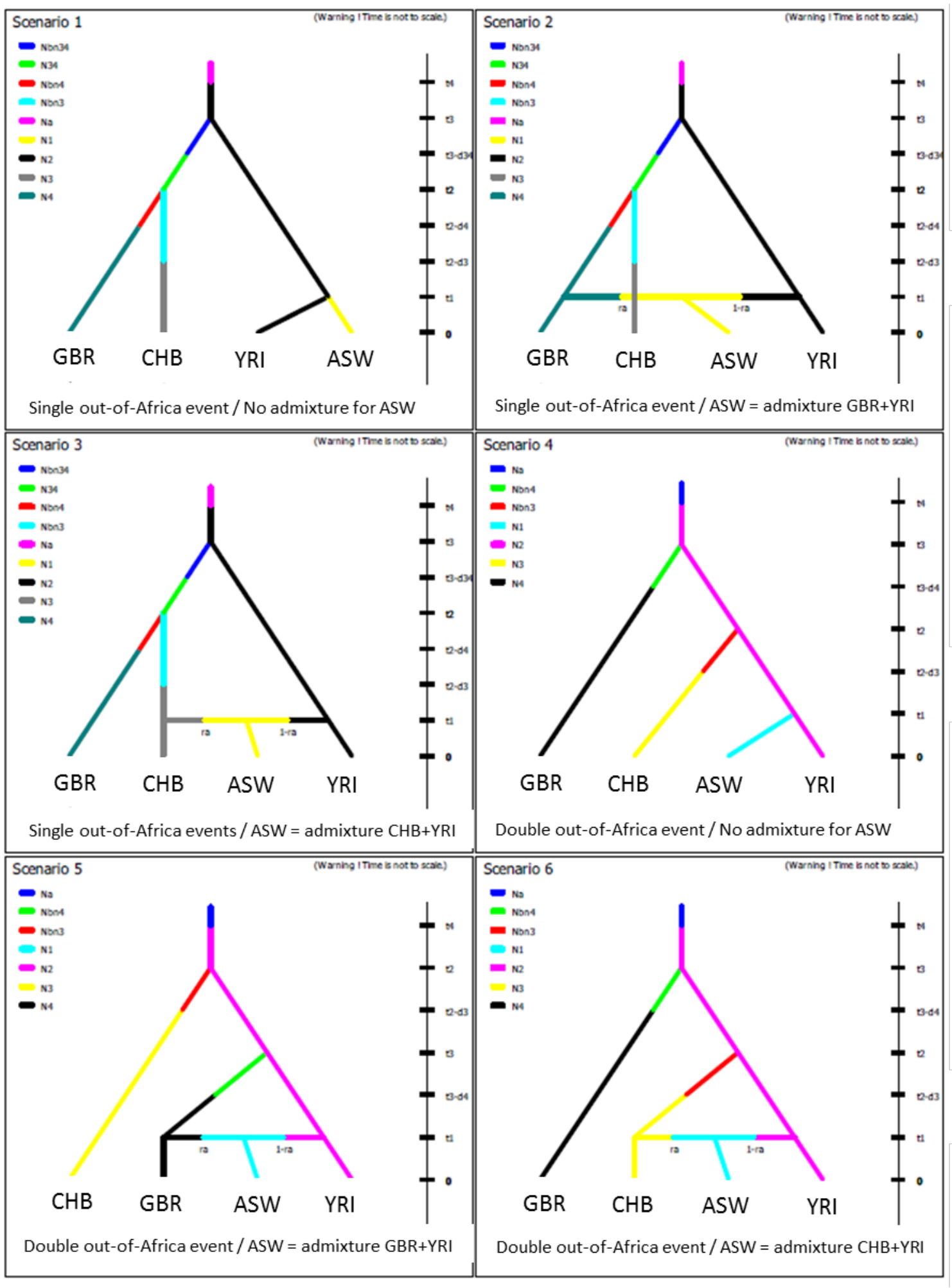}}\par
\caption{{\bf \sffamily \normalsize Six scenarios of evolution of four Human populations genotyped at 50,000
      SNPs.}\label{fig:outofAf} \sffamily \scriptsize The genotyped populations are YRI = Yoruba (Nigeria, Africa), 
      CHB = Han (China, East Asia), GBR = British
      (England and Scotland, Europe), and ASW = Americans of African ancestry (SW USA).  
      The six scenarios differ from each other by one ancient and one recent historical event: (i) a single
      out-of-Africa colonization event giving an ancestral out-of-Africa population which secondarily splits into one
      European and one East Asia population lineage (scenarios 1, 2 and 3), versus two independent out-of-Africa
      colonization events, one giving the European lineage and the other one giving the East Asia lineage
      (scenarios 4, 5 and 6). (ii) The possibility (or not; scenarios 1 and 4) of a recent genetic admixture of ASW
      individuals with their African ancestors and individuals of European (scenarios 2 and 5) or East Asia origins
      (scenarios 3 and 6). The prior distributions of the parameters used to simulate SNP datasets are as followed:
      Uniform[100; 10000] for the split times t2 and t3 (in number of generations), Uniform[1; 30] for the admixture
      (or split) time t1, Uniform[0.05; 0.95] for the admixture rate ra (proportion of genes with a non-African
      origin; only for scenarios with admixture), Uniform[1000; 100000] for the stable effective population sizes
      N1, N2, N4, N4 and N34 (in number of diploid individuals), Uniform[5; 500] for the bottleneck effective
      population sizes Nbn3, Nbn4, and Nbn34, Uniform[5; 500] for the bottleneck durations d3, d4, and d34,
      Uniform[100; 10000] for both the ancestral effective population size Na and the time of change to
      Na. Conditions on time events were t4$>$t3$>$t2 for scenarios 1, 2 and 3, and t4$>$t3 and t4$>$t2 for scenarios 4, 5
      and 6.}
\end{figurehere}

\newpage

\section{Computer software and codes}

For all illustrations based on genetic data, we used the program DIYABC v2.0 \citep{cornuet:etal:2014} to generate the
ABC reference tables including a set of simulation records made of model indices, parameter values and summary
statistics for the associated simulated data. DIYABC v2.0 is a multithreaded program which runs on three operating
systems: GNU/Linux, Microsoft Windows and Apple Mac Os X. Computational procedures are written in C++ and the
graphical user interface is based on PyQt, a Python binding of the Qt framework. The program is freely available to
academic users with a detailed notice document, example projects, and code sources (Linux) from:
\texttt{http://www1.montpellier.inra.fr/CBGP/diyabc}. The reference table generated this way then served as
training database for the random forest constructions. For a given reference table, computations were
performed using the R package \texttt{randomForest} \cite{liaw:wiener:2002}. 
We have implemented all the proposed methodologies in the R package {\sf abcrf} available on the CRAN.

\section{Summary statistics available in the DIYABC software}

\subsection{DIYABC summary statistics on SNP data}

\vspace{0.25cm} Single population statistics \\
\verb+HP0_i+: proportion of monomorphic loci for population i \\
\verb+HM1_i+: mean gene diversity across polymorphic loci \citep{nei:1987} \\
\verb+HV1_i+: variance of gene diversity across polymorphic loci \\
\verb+HMO_i+: mean gene diversity across all loci 

\vspace{0.25cm} \noindent Two population statistics \\
\verb+FP0_i&j+: proportion of loci with null FST distance between the two samples for populations i and j \citep{weir:cockerham:1984} \\ 
\verb+FM1_i&j+: mean across loci of non null FST distances  \\
\verb+FV1_i&j+: variance across loci of non null FST distances  \\
\verb+FMO_i&j+: mean across loci of FST distances \\
\verb+NP0_i&j+: proportion of 1 loci with null Nei's distance  \citep{nei:1972} 
\verb+NM1_i&j+: mean across loci of non null Nei's distances  \\
\verb+NV1_i&j+: variance across loci of non null Nei's distances  \\
\verb+NMO_i&j+: mean across loci of Nei's distances 

\vspace{0.25cm} \noindent Three population statistics \\
\verb+AP0_i_j&k+: proportion of loci with null admixture estimate when pop. i comes from an admixture between j and k \\
\verb+AM1_i_j&k+: mean across loci of non null admixture estimate \\
\verb+AV1_i_j&k+: variance across loci of non null admixture estimated \\
\verb+AMO_i_j&k+: mean across all locus admixture estimates 

\subsection{DIYABC summary statistics on microsatellite data}

\vspace{0.25cm} Single population statistics \\
\verb+NAL_i+: mean number of alleles across loci for population i \\
\verb+HET_i+: mean gene diversity across loci \citep{nei:1987} \\
\verb+VAR_i+: mean allele size variance across loci \\
\verb+MGW_i+: mean M index across loci \citep{garza:williamson:2001,excoffier:etal:2005} 

\vspace{0.25cm} \noindent Two population statistics \\ 
\verb+N2P_i&j+: mean number of alleles across loci for populations i and j \\
\verb+H2P_i&j+: mean gene diversity across loci  \\
\verb+V2P_i&j+: mean allele size variance across loci \\
\verb+FST_i&j+: FST \citep{weir:cockerham:1984} \\
\verb+LIK_i&j+: mean index of classification \citep{rannala:mountain:1997,pascual:etal:2007} \\
\verb+DAS_i&j+: shared allele distance \citep{chakrabortry:jin:1993} \\
\verb+DM2_i&j+: $(\delta\mu)^2$ distance \citep{goldstein:etal:1995} 

\vspace{0.25cm} \noindent Three population statistics \\
\verb+AML_i_j&k+: Maximum likelihood coefficient of admixture when pop. i comes from an admixture between j and k \citep{choisy:etal:2004}

\newpage


\end{document}